\documentclass[%
 reprint,
 amsmath,amssymb,
 aps,
]{revtex4-2}

\usepackage{graphicx}
\usepackage{dcolumn}
\usepackage{bm}
\usepackage{xcolor}
\usepackage[colorlinks=true, citecolor=blue]{hyperref}
\usepackage[linesnumbered,ruled,vlined]{algorithm2e}
\usepackage{algpseudocode}

\usepackage{bbding}
\usepackage{booktabs}
\usepackage{subfigure}

\newtheorem{definition}{Definition}
\newtheorem{proposition}{Proposition}

\begin{document}

\preprint{APS/123-QED}

\title{Ultra-Early Prediction of Tipping Points: Integrating Dynamical Measures with Reservoir Computing}

\author{Xin Li$^{1,2}$, Qunxi Zhu$^{2,3,*}$, Chengli Zhao$^{1,*}$, Bolin Zhao$^{2,3}$, Xue Zhang$^{1}$, Xiaojun Duan$^{1}$, and Wei Lin$^{2,3,4}$}
\email{E-mail: qxzhu@fudan.edu.cn, chenglizhao@nudt.edu.cn, or wlin@fudan.edu.cn}
\affiliation{$^1$College of Science, National University of Defense Technology, Changsha, Hunan 410073, China\\ $^2$Research Institute of Intelligent Complex Systems and MOE Frontiers Center for Brain Science, Fudan University, Shanghai 200433, China \\ $^3$School of Mathematical Sciences, SCMS, SCAM, and CCSB, Fudan University, Shanghai 200433, China \\ $^4$Shanghai Artificial Intelligence Laboratory, Shanghai 200232, China.}


\begin{abstract}
Complex dynamical systems—such as climate, ecosystems, and economics—can undergo catastrophic and potentially irreversible regime changes, often triggered by environmental parameter drift and stochastic disturbances. These critical thresholds, known as tipping points, pose a prediction problem of both theoretical and practical significance, yet remain largely unresolved. To address this, we articulate a model-free framework that integrates the measures characterizing the stability and sensitivity of dynamical systems with the reservoir computing (RC), a lightweight machine learning technique, using only observational time series data. The framework consists of two stages. The first stage involves using RC to robustly learn local complex dynamics from observational data segmented into windows. The second stage focuses on accurately detecting early warning signals of tipping points by analyzing the learned autonomous RC dynamics through dynamical measures, including the dominant eigenvalue of the Jacobian matrix, the maximum Floquet multiplier, and the maximum Lyapunov exponent. Furthermore, when these dynamical measures exhibit trend-like patterns, their extrapolation enables ultra-early prediction of tipping points significantly prior to the occurrence of critical transitions. We conduct a rigorous theoretical analysis of the proposed method and perform extensive numerical evaluations on a series of representative synthetic systems and eight real-world datasets, as well as quantitatively predict the tipping time of the Atlantic Meridional Overturning Circulation system. Experimental results demonstrate that our framework exhibits advantages over the baselines in comprehensive evaluations, particularly in terms of dynamical interpretability, prediction stability and robustness, and ultra-early prediction capability.

\end{abstract}

\maketitle


\section{Introduction}
Complex dynamical systems (CDSs), ranging from the biological systems \cite{Lesterhuis2017DynamicVS} and the development of psychiatric disorders \cite{van2014critical} to climate \cite{clark2002role}, economic \cite{May2008ComplexSE}, ecological \cite{Hastings2018TransientPI}, and social systems \cite{centola2018experimental}, can often experience sudden transitions from one state to drastically different states when subjected to minor changes in environmental conditions. The locus of these transitions are commonly known as tipping points, where states with undesirable properties always emerge. For instance, in climate systems, crossing tipping points often has severe consequences for human society and even exacerbate global warming \cite{lenton2019climate}. Thus, the accurate prediction of tipping points in CDSs is of critical importance and broad interest.

Recent advancements in both theoretical and empirical research have substantially enhanced our ability to identify tipping points in CDSs. As a system nears a tipping point, it frequently exhibits a phenomenon known as critical slowing down (CSD) \cite{Dakos2012RobustnessOV}, which reflects a reduced capacity to recover from local disturbances \cite{Wissel1984AUL,vanNes2007SlowRF}. Building on this insight, Scheffer et al. \cite{Scheffer2009EarlywarningSF} synthesized the theories of bifurcation and CSD to identify tipping points using classical early warning signals (EWSs), such as Lag-1 autocorrelation \cite{Dakos2008SlowingDA}, variance \cite{Carpenter2006RisingVA}, skewness, and flickering \cite{Guttal2008ChangingSA}. While these classical EWSs provide valuable qualitative indicators of critical transitions, their performance can be inconsistent across different systems, making it difficult to establish accurate thresholds for tipping point identification or even its early warning.

To address these challenges, Bury et al. \cite{Bury2021DeepLF} introduced a deep learning (DL) model trained on the synthetic bifurcation data, enabling it to predict tipping points in complex scenarios. Building on this, Grziwotz et al. \cite{Grziwotz2023AnticipatingTO} proposed a new EWS based on dynamical eigenvalue (DEV) analysis, which leverages the bifurcation theory and local linearization methods to predict tipping points more effectively across both simulated and real-world datasets. Despite their promise, both approaches have limitations. The DL method is often regarded as a ``black box'' with limited interpretability and minimal incorporation of prior knowledge from dynamical systems theory. Conversely, the DEV method, while more theoretically transparent, lacks the adaptability of neural networks for capturing nonlinear dynamics and is restricted to bifurcations of fixed-point attractors, rendering them unsuitable for periodic or even chaotic systems.

Reservoir computing (RC) \cite{Lukoeviius2009ReservoirCA,Jaeger2001TheechoST,li2024higher}, a lightweight machine learning technique, has gained increasing attention for tasks such as reconstruction and prediction in CDSs \cite{pathak2018model,zhu2019detecting,Tang2020IntroductionTF,Platt2022ASE,Li2023TippingPD}. RC operates at the edge of stability or chaos \cite{Ascoli2022EdgeOC,Carroll2020DoRC}, allowing for the emergence of diverse dynamical behaviors in its hidden states. This flexibility makes RC particularly well-suited to modeling nonlinear dynamics, such as equilibria, limit cycles, and chaotic attractors \cite{Gilpin2021ChaosAA,flynn2021multifunctionality} solely from the observational time series data. Further improvements have been made by incorporating system parameters into RC methods, as demonstrated by Kong et al. \cite{kong2021machine},  Patel et al. \cite{Patel2022UsingML} and Panahi et al. \cite{panahi2024adaptable}.  These enhancements allow for more accurate modeling of dynamical systems. While RC has been successfully applied to tipping point prediction \cite{kong2021machine,Patel2022UsingML,panahi2024adaptable}, such successes typically depend on access to comprehensive system parameters and related information, which are often difficult to acquire in real-world scenarios.

To overcome these difficulties, we articulate a data-driven framework called the RC-based dynamical measure (RCDyM) method, which integrates RC techniques with dynamical measures. Notably, the proposed RCDyM method focuses on estimating three main dynamical measures related to stability and sensitivity under local perturbations. Consequently, our framework does not require varying system parameters as inputs, nor does it depend on precise global modeling of the dynamics at tipping points. This generality makes our framework more versatile, enabling it to be applicable to a broader range of real-world CDSs.

The remaining of this article is thus organized as follows.  We first introduce the framework of the RCDyM method.   This framework, by employing a sliding window strategy, firstly trains the RC, thereby allowing for the analysis of the high-dimensional \textit{autonomous} RC to estimate the dynamical indicators of previously unknown systems. Subsequently, these dynamical metrics can not only serve as early-warning signals for critical transitions, but also enable ultra-early prediction of tipping points by fitting and extrapolating their evolutionary trends. Then, we demonstrate the RCDyM method across various scenarios, showing its outperformance over the state-of-the-art (SOTA) baseline methods.  Indeed, we show that the RCDyM method can predict critical transitions associated with equilibria, periodic cycles, and chaotic dynamics. Additionally, we also show that the RCDyM method can predict the occurrence and types of bifurcations in real-world CDSs, further highlighting its broad applicability and practical relevance.

\section*{Results} 
\subsection*{The Framework of RCDyM Method}
\subsubsection*{Continuous Reservoir Computing.}
We consider a nonlinear CDS comprising $N$ variables of the following general form,
\begin{equation}\label{E_orig}
	\dot{\bm{s}}(t) = \bm{f}[\bm{s}(t),\bm{p}(t),t],
\end{equation}
where $\bm{s}(t)=[s_1(t),...,s_N(t)]^{\top}$ is the $N$-dimensional ($N$-D) state of the nonlinear system at time $t$, $\bm{p}(t)=[p_1(t),...,p_M(t)]^{\top}$ denotes the $M$-D time-varying parameter vector, which is assumed to change slowly over time unless otherwise specified, and $\bm{f}[\bm{s}(t),\bm{p}(t),t]=\{{f}_1[\bm{s}(t),\bm{p}(t),t],...,{f}_N[\bm{s}(t),\bm{p}(t),t]\}^{\top}$ represents the time-varying nonlinear vector field with the external input (parameter vector) $\bm{p}(t)$. 

\begin{figure*}[t]
	\centering
	\includegraphics[width=1.0 \textwidth]{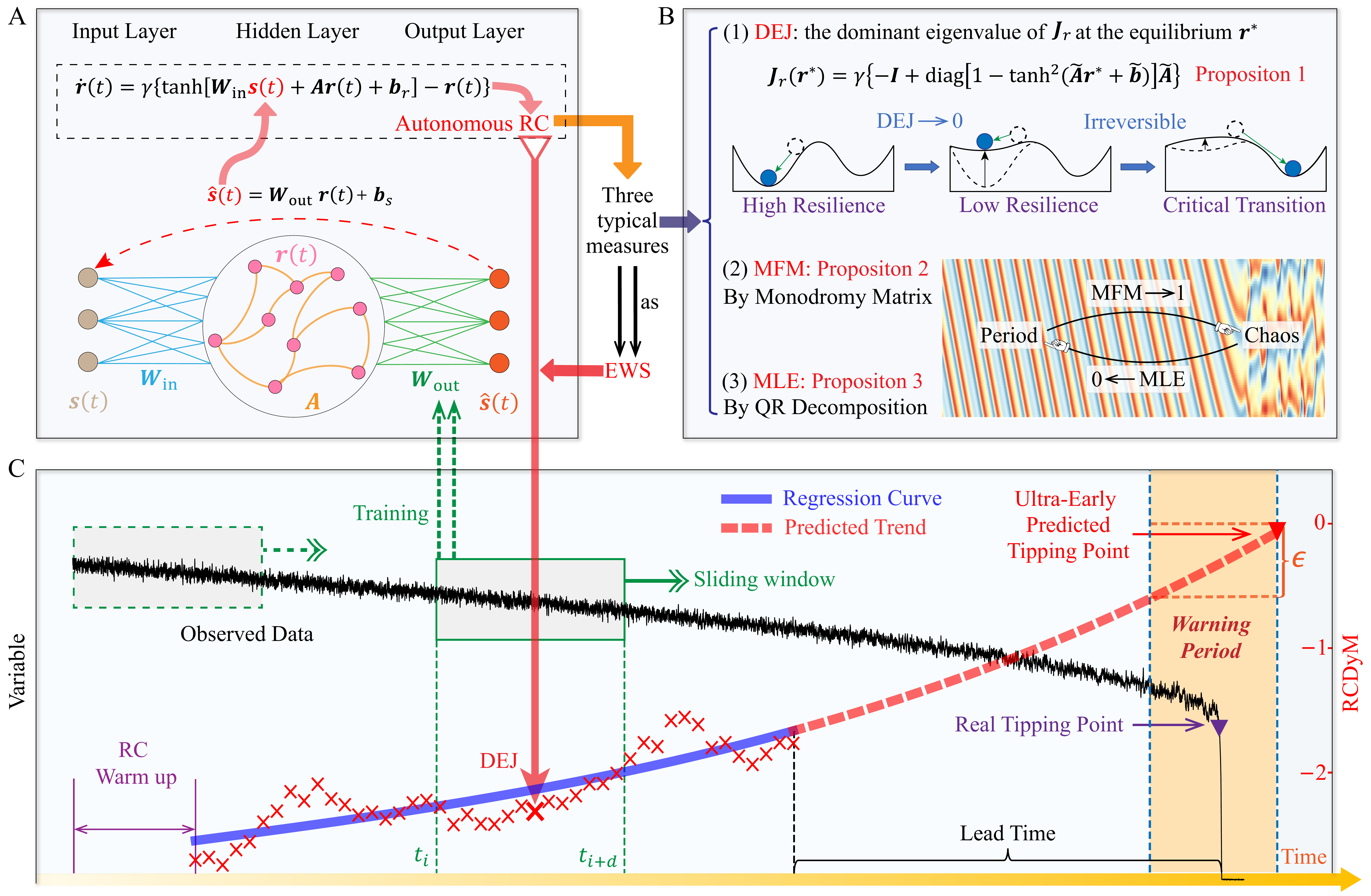}
	\vspace{-0.1cm}
	\subfigure{\label{fg:modelA}} 
	\subfigure{\label{fg:modelB}} 
	\subfigure{\label{fg:modelC}}
	\caption{\textbf{Schematic diagram for the proposed RCDyM method.} ($A$) The basic structure of  RC consists of the input layer, the hidden layer, and the output layer. The continuous-time, autonomous RC is obtained by replacing the input with the predicted state using the learned RC. ($B$)  Three RCDyMs are depicted with the application scenarios for the tipping point prediction. ($C$) An illustrative example of tipping point prediction using the DEJ measure, one of the RCDyMs: Utilizing trend analysis for ultra-early prediction of tipping points.} 
	\label{F_model}
	\vspace{-0.5cm}
\end{figure*}

As illustrated in Fig.~\ref{fg:modelA}, the classical RC framework typically consists of three main components: the input layer, the hidden layer (reservoir), and the output layer. Initially, the input layer embeds the observational data $\bm{s}(t)$ of the system \eqref{E_orig} into a higher-dimensional reservoir network via the input matrix $W_\text{in}$. Then, the hidden state $\bm{r}(t)$ within the reservoir network evolves according to the following continuous-time, controlled dynamics with the input $\bm{s}(t)$:
\begin{equation} \label{contRC}	
	\dot{\bm{r}}(t) = \gamma\{-\bm{r}(t) + \tanh[\bm{A}\bm{r}(t) + \bm{W}_{\text{in}} \bm{s}(t) + \bm{b}_r]\},
\end{equation}
where $\gamma$ is the time scale constant, $\bm{A}$ represents the adjacency matrix of the dimension $n \times n$, $\bm{W}_{\text{in}}$ denotes the input matrix of the dimension $n \times N$, and $\bm{b}_r$ is the dynamical bias term with the dimension $n \times 1$. It is important to note that the matrices $\bm{A}$ and $\bm{W}_{\text{in}}$ are randomly generated and remain fixed throughout the training process. The only trained module is the output layer, typically a linear transformation via the output matrix $\bm{W}_{\text{out}}$ of the dimension $N \times n$ along with an $N$-D bias term $\bm{b}_s$, which maps the hidden state space back to the original state space:
\begin{equation} \label{output_layer}	
	\hat{\bm{s}}(t) = \bm{W}_{\text{out}}\bm{r}(t) + \bm{b}_s.
\end{equation}
Notably, the output layer can be trained using the ridge regression, yielding the closed form of the $\bm{W}_{\text{out}}$ and $\bm{b}_s$. During this training process, the loss function:
\begin{equation}\label{Eloss}
	\mathcal{L} = \int_{t} \left\|\hat{\bm{s}}(t)-\bm{s}(t)\right\|^2 \text{d}t + \lambda (\|\bm{W}_{\text{out}}\|^2+\|\bm{b}_s\|^2)
\end{equation}
is minimized.  Here,  $\|\cdot\|$ denotes the standard $2$-norm, corresponding to the $l_2$ regularization, and $\lambda$ is a positive regularization coefficient. In practice, the sum of predictive errors at discrete observational points is used as an approximation for the integral term in the loss function \eqref{Eloss}. Moreover, additional details on classical RC technique can be found in Section 2.A of the Supporting Information.

After training, the state of the underlying CDS can be predicted using the Eqs.~\eqref{contRC} and \eqref{output_layer}.  By substituting $\bm{s}$ with $\hat{\bm{s}}$ in the controlled RC \eqref{contRC}, an autonomous high-dimensional reservoir dynamical system is obtained, described by the following autonomous dynamical system:
\begin{equation}\label{autoRC}
	\dot{\bm{r}}(t) = \gamma \{-\bm{r}(t) + \tanh[\tilde{\bm{A}}\bm{r}(t)+\tilde{\bm{b}}] \},
\end{equation}
where, for notational simplicity, $\tilde{\bm{A}} := \bm{A}+\bm{W}_{\text{in}} \bm{W}_{\text{out}}$ and $\tilde{\bm{b}} := \bm{W}_{\text{in}}\bm{b}_s+\bm{b}_r$.

\subsubsection*{Three Typical Dynamical Measures}
As shown in Fig.~\ref{fg:modelB}, to predict the critical transition, we use three typical dynamical measures for analyzing RC in the autonomous  system \eqref{autoRC}. 

First, the dominant eigenvalue of the Jacobian matrix (DEJ) at a system's equilibrium is a representative measure used in the bifurcation analysis, providing theoretical insights for EWS of critical transitions. Although the DEJ of system~\eqref{E_orig} cannot be directly obtained from the observational data, our analytical arguments demonstrate that the DEJ of the RC dynamics, as described in system~\eqref{autoRC}, offers a robust approximation of the DEJ for the original system (see Proposition \ref{P_DEJ}). Consequently, the stability of the original system can be inferred by evaluating the DEJ of system \eqref{autoRC} at its equilibrium. To locate the equilibrium of the high-dimensional system \eqref{autoRC}, we apply the multivariate Newton’s method based on the estimated Jacobian matrix, utilizing the Python's ``fsolve'' function (detailed in Methods section).

\begin{proposition}\label{P_DEJ} 
	Denote by $\bm{J}_r(\bm{r}^*)$ the Jacobian matrix of  system \eqref{autoRC} at the equilibrium $\bm{r}^*$: 
	\begin{equation}\label{Jacobi}
		\bm{J}_r(\bm{r}^*) = \gamma \{-\bm{I} + \text{\rm diag}[1-\tanh^2(\tilde{\bm{A}}\bm{r}^* + \tilde{\bm{b}})]\tilde{\bm{A}}\}.
	\end{equation}
	Suppose that the state $\bm{s}$ of system \eqref{E_orig} satisfies $\bm{s} = \bm{W}_{\text{\rm out}} \bm{r} + \bm{b}_s$, and that the dominant eigenvector $\bm{v}_1$ of $\bm{J}_r$ is not in the null space of $\bm{W}_\text{\rm out}$, i.e., $\bm{W}_\text{\rm out}\bm{v_1}\ne 0$. Then, the dominant eigenvalue $\lambda_{\text{\rm DEJ}}$ of $\bm{J}_r$ is equal to $\hat{\lambda}_{\text{DEJ}}$, where $\hat{\lambda}_{\text{DEJ}}$ is the DEJ of the original system \eqref{E_orig}. 
\end{proposition}

Additionally, the maximum Floquet multiplier (MFM) \cite{barone1977floquet} is another crucial measure for evaluating the stability of the periodic orbits. Similar to the DEJ, the stability of periodic orbits in the original system can be estimated by calculating the MFM of the high-dimensional dynamics in system~\eqref{autoRC}. Here, we propose an automatic method for detecting periodicity based on the autocorrelation coefficient and error analysis (detailed in Methods section). Furthermore, the maximum Lyapunov exponent (MLE) is an established measure for quantifying any orbit including chaotic dynamics of a nonlinear system.   Actually, the MLE of system \eqref{autoRC} can be effectively used to estimate the MLE of the original system \eqref{E_orig}, a relationship that has been empirically validated in earlier studies \cite{pathak2017using}. Building on Proposition \ref{P_DEJ}, we present the following Propositions \ref{P_MFM} and \ref{P_MLE} for estimating the MFM and the MLE, respectively, of the original system~\eqref{E_orig} by analyzing the high-dimensional RC in \eqref{autoRC}.  Complete arguments for demonstrating these propositions are provided in Supporting Information Section 1.

\begin{proposition}\label{P_MFM}
	Denote by $T_\text{p}$ the period of system \eqref{E_orig} and 
	by $\bm{\Phi}(T_\text{p})$ the Monodromy matrix of system \eqref{autoRC}.  Here,  the matrix $\bm{\Phi}$  obeys the following differential equation:
	$$\dot{\bm{\Phi}}(t) = \bm{J}_r[\bm{r}(t)]\bm{\Phi}(t), \quad \bm{\Phi}(0) = \bm{I},
	$$
	where $\bm{J}_r$ is the Jacobian matrix defined in \eqref{Jacobi} but along the trajectory $\bm{r}(t)$.
	If the state $\bm{s}$ of system \eqref{E_orig} satisfies $\bm{s} = \bm{W}_{\text{\rm out}} \bm{r} + \bm{b}_s$, and the dominant eigenvector $\bm{v}_1$ of $\bm{J}_r$ is not in the null space of $\bm{W}_\text{\rm out}$, i.e., $\bm{W}_\text{\rm out}\bm{v_1}\ne 0$, then the dominant eigenvalue $\lambda_{\text{\rm MFM}}$ of $\bm{\Phi}(T_\text{p})$ is equal to $\hat{\lambda}_{\text{\rm MFM}}$, where $\hat{\lambda}_{\text{\rm MFM}}$ is the MFM of the original system \eqref{E_orig}. 
\end{proposition}

\begin{proposition}\label{P_MLE}
	Denote by $\lambda_{\text{\rm MLE}}$ the MLE of system \eqref{autoRC}.  Here, we use the QR decomposition method \cite{abarbanel2012analysis} to compute $\lambda_{\text{\rm MLE}}$ (detailed in Methods section). If the state $\bm{s}$ of system \eqref{E_orig} satisfies $\bm{s} = \bm{W}_{\text{\rm out}} \bm{r} + \bm{b}_s$, and the dominant eigenvector $\bm{v}_1$ of $\bm{J}_r$ is not in the null space of $\bm{W}_\text{\rm out}$, i.e., $\bm{W}_\text{\rm out}\bm{v_1}\ne 0$, then $\lambda_{\text{\rm MLE}} = \hat{\lambda}_{\text{\rm MLE}}$, where $\hat{\lambda}_{\text{\rm MLE}}$ is the MLE of system \eqref{E_orig}. 
\end{proposition}

\subsubsection*{Ultra-Early Prediction of Tipping Points}
To achieve online and ultra-early prediction of tipping points, we adopt a sliding window strategy for the real-time computation of dynamical measures. As illustrated in Fig.~\ref{fg:modelC}, we utilize a data window of length $d$ in each iteration to estimate the output matrix $\bm{W}_{\text{out}}$, and then we obtain the corresponding autonomous RC in \eqref{autoRC}. Then, we compute dynamical measures based on the autonomous RC and utilize each of them as an EWS. The timestamp for the measure is assigned based on the midpoint time within the window, i.e., the timestamp for the $i$-th measure is $\frac{1}{2}(t_i + t_{i+d})$. By employing a sliding window with a step length of $k$ during each iteration, our method generates time-varying dynamical measures (red scatter points in Fig.~\ref{fg:modelC}). Subsequently, we perform a regression analysis to fit the trend of the measures, as depicted by the blue solid line and the red dashed line in Fig.~\ref{fg:modelC}. In our work, we employ refined polynomial fitting of a relatively lower order to capture the trend of the measures (see Supporting Information Section 2.D for details). Clearly, the estimated measures can themselves serve as novel EWSs for early warning. When these measures exhibit trend-like patterns that are readily fittable, the RCDyM method enables quantitative prediction well before the onset of critical transitions, thereby facilitating ultra-early prediction, and the strict definition of ultra-early prediction is provided in Definition \ref{ultra}.

Specifically, we introduce a positive threshold $\epsilon$ determined by noise disturbances and other related factors. For the DEJ measure, if the real part of the DEJ at an equilibrium increases beyond $-\epsilon$, it indicates that the system may tend to lose its resilience and is prone to significant and potentially irreversible shifts (see the warning period in Fig.~\ref{fg:modelC}). Notably, critical transitions are not always induced prematurely by noise or other factors. Thus, the time at which the DEJ curve reaches the critical threshold 0 remains the most fundamental prediction target, as shown by the red triangle in Fig.~\ref{fg:modelC}. Analogously, for the MFM measure, if the real part exceeds $1 - \epsilon$, the system's periodic orbit is considered to be near a tipping point. And, if the MLE in a chaotic system decreases below $\epsilon$, this suggests a potential transition from chaos to other ordered dynamical behaviors. 

In addition, as a complement to the current version of the continuous-time RCDyM method, we also provide a discrete-time version in Supporting Information Section 2.B, and discuss the relationship between these two versions in Supporting Information Section 2.C. Moreover, the ability of the RC model to capture dynamics ensures that the choice of hyperparameters aligns with those typically used in conventional RC forecasting tasks. Therefore, we can further refine the selection of appropriate RC hyperparameters guided by the predictive performance of reservoir computing (see Methods section and Supporting Information Section 4 for more details). To validate the effectiveness and robustness of our framework, we first conduct comparative analysis between the RCDyM method and those baseline methods in Table \ref{T_compare}, and then we perform a series of experiments on datasets from both simulated and real-world systems.

\subsection*{Validation in Classical Dynamical Systems}
\subsubsection*{Classical Bifurcation Models}
We begin by validating our method on four different, well-established bifurcation systems, i.e., fold, period-doubling, pitchfork, and Hopf bifurcations. The corresponding dynamical equations and bifurcation points for these models are presented in Table \ref{T_bifur}. To increase the realism of the experimental setup, we introduce a time-varying parameter $p(t)$, which traverses the bifurcation point of the parameter. Furthermore, as shown in Table \ref{T_bifur}, we incorporate the Gaussian white noise into the state variable $s(t)$ in continuous-time systems, with the noise being zero mean and deviation $\sigma_\text{n} = \omega \zeta s$, where $\omega$ represents the noise intensity, set to a default value of $0.01$, and $\zeta$ is a random variable drawn from a standard normal distribution.

\begin{figure*}[t]
	\centering
	\includegraphics[width=1.0\linewidth]{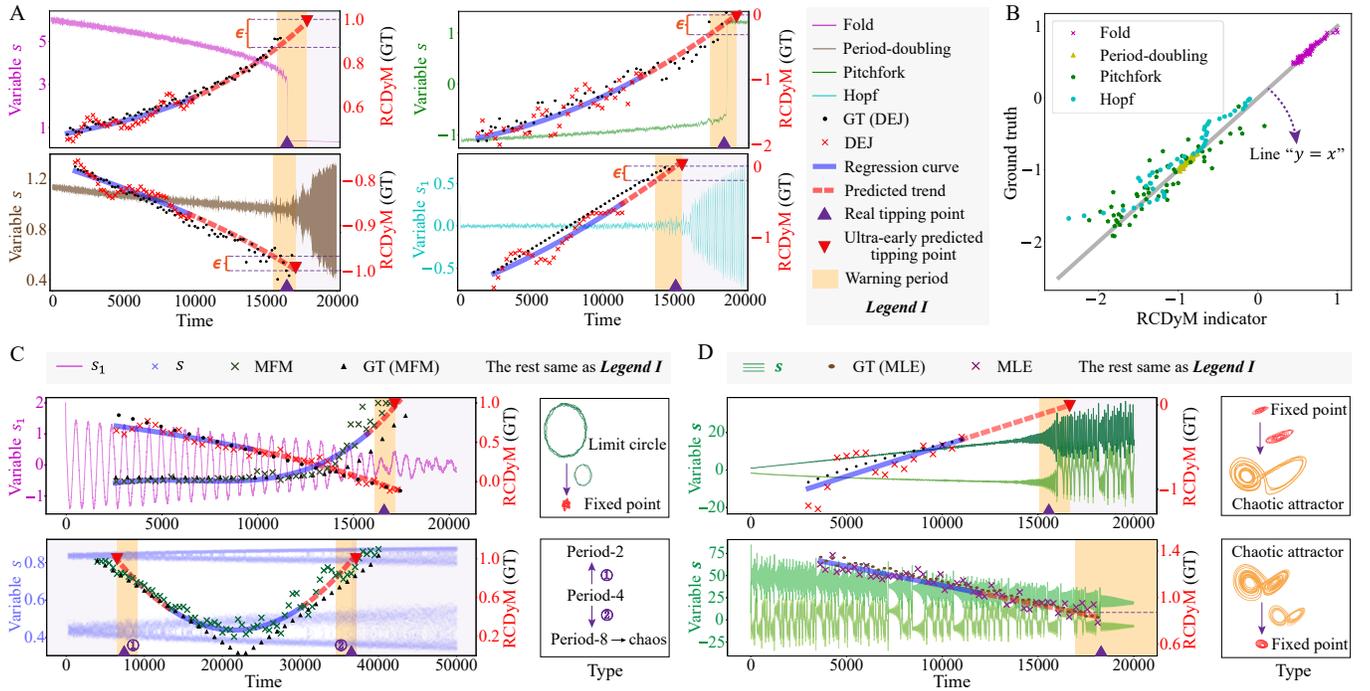}
	\subfigure{\label{fg:bifurA}} \subfigure{\label{fg:bifurB}}
	\subfigure{\label{fg:bifurC}} \subfigure{\label{fg:bifurD}}
	\vspace{-0.5cm}
	\caption{\textbf{Tipping points prediction for representative continuous/discrete-time dynamical systems, including four bifurcation systems and two chaotic systems.} ($A$) Ultra-early prediction of tipping points in systems with different bifurcations, viz., fold, period-doubling, pitchfork, and Hopf bifurcations. The time interval $\Delta t$ was set to 0.1 for pitchfork bifurcation and 0.05 for Hopf bifurcation. For each experiment, the early-warning parameter $\epsilon$ is taken as 15\% of the total DEJ measure range. ($B$) The estimated DEJ measures using the RCDyM method and the ground truth (GT) calculated directly using the original system equations are distributed around the line $y=x$. ($C$) Predicting the critical transition from a limit cycle to an equilibrium in the Hopf bifurcation system, as well as the critical transitions from period-4 to period-2 and period-8 dynamics in the system governed by the Logistic map. ($D$) Predicting the critical transitions from an equilibrium to chaos and from chaos to an equilibrium in the Lorenz63 system.}
	\label{F_bifur}
	\vspace{-0.1cm}
\end{figure*}

For the discrete bifurcation systems, we generate experimental time series with $T = 20,000$ data points, where the time-varying parameter $p_i$ is defined as $p_i = \frac{ki}{T} + b$, for $i= 0,1,...,T-1$. Similarly, for the continuous systems, we sample $T$ data points at equidistant intervals of $\Delta t$, with the time-varying parameter set as $p(t) = \frac{kt}{T\Delta t} + b$. To ensure the uniformity in exposition, the indices of the data points are consistently used as the proxies for time points. The parameter pair $(k, b)$ is set as follows for different bifurcation experiment: $(1, 1)$ for the fold bifurcation, $(1.65, -0.50)$ for the pitchfork bifurcation, $(0.15, 0.25)$ for the period-doubling bifurcation, and $(-2, 2.5)$ for the Hopf bifurcation. 

Next, we apply the discrete-time RCDyM method to compute the DEJ measures for the experiments of the fold and the period-doubling bifurcations, and the continuous-time RCDyM for the experiments of the pitchfork and the Hopf bifurcations. The experimental results are displayed in Fig.~\ref{fg:bifurA}, utilizing dual $y$-axis plots. In these plots, the left $y$-axis represents the variable $\bm{s}$, while the right $y$-axis shows the DEJ measure alongside the corresponding ground truth (GT). It is evident that our method effectively provides ultra-early predictions of the tipping points. The ground-truth DEJ values, computed through the actual system equations, are depicted as black dots in Fig.~\ref{fg:bifurA}, and the correlation between the RCDyM and the GT is shown in Fig.\ref{fg:bifurB}.   Clearly, the estimated DEJ follows a growth trend closely aligned with the GT, and both approach the tipping point almost simultaneously. This indicates that our method successfully captures the underlying dynamical information, even in the presence of a certain strength of noise. Furthermore, in some dynamical systems, ultra-early predicted tipping points may exhibit a slight temporal lag relative to the real tipping points, primarily due to noise-induced premature transitions prior to the bifurcation parameter threshold. By calibrating the value of $\epsilon$ to align with the noise intensity, we accurately map the warning period for true tipping events.	To further investigate the performance of RCDyM in ultra-early prediction tasks, we conduct a supplementary experimental analysis in Supporting Information Section 5.C.

Our method can further identify several classical types of bifurcations. Specifically, as shown in the period-doubling bifurcation experiment in Fig.~\ref{fg:bifurA}, the imaginary part of the DEJ, denoted as Im(DEJ), is zero, while the real part of the DEJ, denoted as Re(DEJ), converges to $-1$, which is the typical feature of a period-doubling bifurcation. In the Hopf bifurcation experiment, Re(DEJ) approaches $0$, while Im(DEJ) remains nonzero. In the fold and the pitchfork bifurcation experiments, for the discrete-time system, Re(DEJ) converges to $1$ and Im(DEJ) equals zero. And for the continuous-time system, Re(DEJ) tends toward $0$ and Im(DEJ) remains zero. Detailed discussion and experimental results are provided in Section 5.B of the Supporting Information. The results show that the RCDyM method can effectively track both the real and imaginary parts of the dominant eigenvalue over time, allowing for the identification of several fundamental bifurcation types.


\subsubsection*{Periodic and Chaotic Systems}
We begin by investigating the critical transition from the limit cycle to the equilibrium in the Hopf bifurcation system, where $p(t) = -2.3 \times \frac{t}{T\Delta t} + 2$. Using the continuous-time RCDyM method, we compute both the DEJ and the MFM, as shown in the upper part of Fig.~\ref{fg:bifurC}. The results demonstrate that the MFM measure effectively estimates the stability of periodic orbits. As the system approaches the critical transition, the modulus of the MFM measure converges to $1$, thereby facilitating the prediction of the tipping point. Furthermore, the DEJ measure gradually transitions from positive values toward zero, indicating that our method successfully captures the process in which the potential equilibrium point shifts from instability to stability, thus enabling the prediction of critical transitions. 

Next, we consider a discrete-time system governed by the Logistic map, as shown in Eq.~\eqref{logistic_map}.  Here, we set $p_i = \frac{0.13 i}{T} + 3.44$ with $T = 50,000$ and $i = 0, 1, ..., T-1$ to generate the time series data and compute the MFM measure using the discrete-time RCDyM method. As illustrated in the lower part of Fig.~\ref{fg:bifurC}, our method effectively predicts the critical transitions, including the bifurcation from period-$4$ to period-$2$ dynamics (on the left side) and the transition to chaos (on the right side).  Strictly speaking, as the parameter $p$ increases, the system should initially undergo a transition from period-$4$ to period-$8$ dynamics, followed by a gradual shift to period-$16$ dynamics, eventually leading to chaos. However, since the transition from period-$8$ to chaos occurs over a very short range (less than the window length $d$), we thus approximate this as a direct transition from period-$4$ to chaos.

We now turn our attention to a more complex chaotic Lorenz63 system \cite{lorenz1963chaos}, see system in~\eqref{E_lorenz63}. As shown in the upper part of Fig.~\ref{fg:bifurD}, we set $p(t) = 22 \times \frac{t}{T \Delta t} + 2$, $\Delta t = 0.01$, and $\omega = 0.001$ to generate the experimental data, capturing the transition from an equilibrium to chaos. It is evident that our method effectively captures EWSs preceding the tipping point. Additionally, we set $p(t) = -30 \times \frac{t}{T \Delta t} + 50$ to generate the experimental data representing the transition from chaos to an equilibrium in a reverse direction, as illustrated in the lower part of Fig.~\ref{fg:bifurD}. The results indicate that our method accurately estimates the changes in the MLE of the system as the time-varying parameter $p(t)$ approaches the critical threshold, aligning closely with the GT (i.e., using the MLE estimates based on the original equations). At the critical threshold, the true MLE exhibits a sudden transition from a positive value to zero, reflecting the inherent dynamical properties of system \eqref{E_lorenz63}. Nevertheless, we observe a reduction in the level of chaos before the tipping point, suggesting the need of selecting a larger warning threshold $\epsilon$ (e.g., 0.8) to anticipate such tipping events in advance. This phenomenon is common in critical transitions from chaos to other stationary behavior.

\begin{figure}[t]
	\centering
	\includegraphics[width=1.0\linewidth]{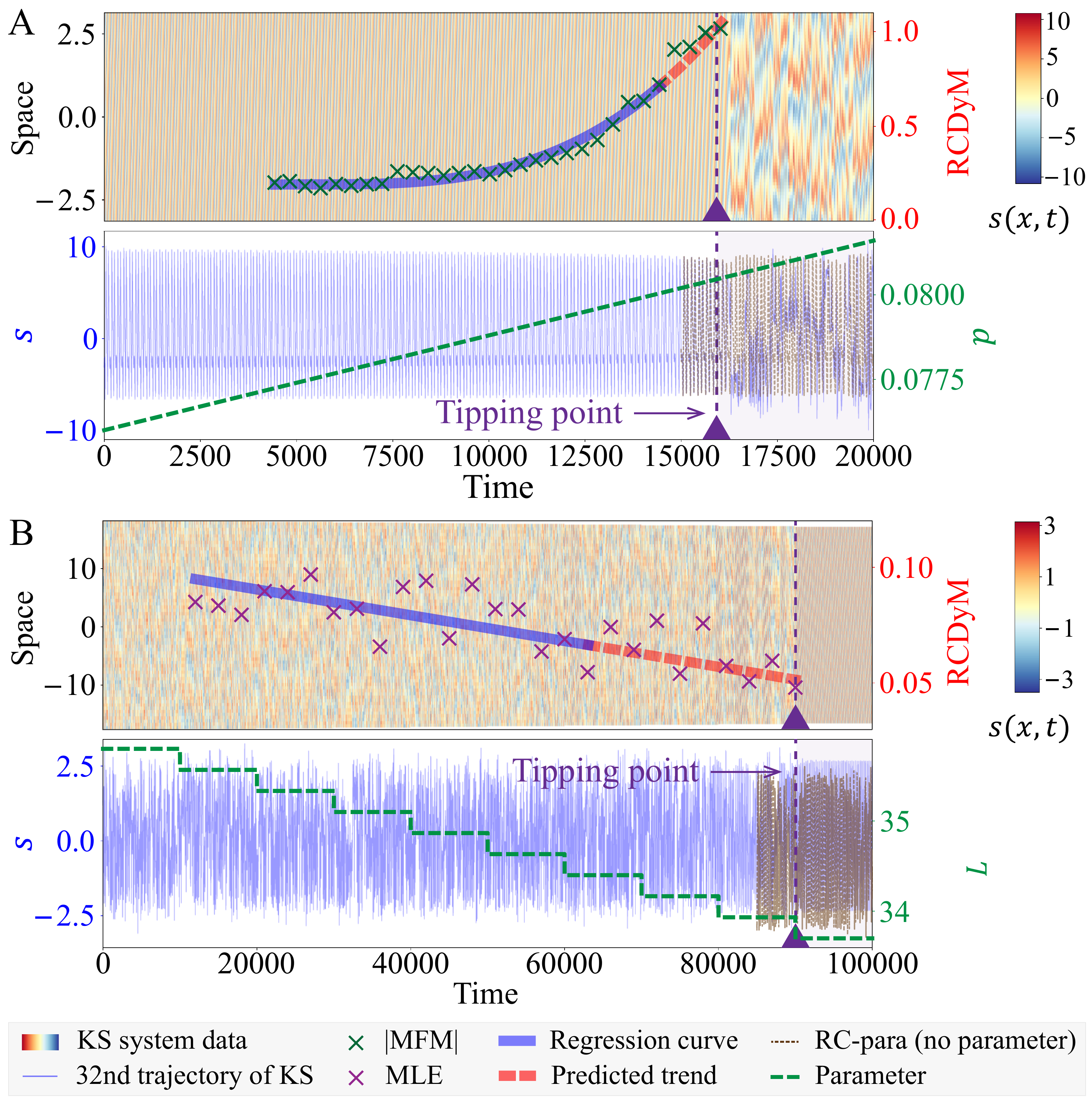}
	\subfigure{\label{fg:KSA}} 
	\subfigure{\label{fg:KSB}}
	\vspace{-0.6cm}
	\caption{\textbf{Tipping points prediction for the KS system.} ($A$) and ($B$) illustrate the predicted transitions in the Kuramoto-Sivashinsky (KS) system: from periodic behavior to chaotic behavior, and from chaotic behavior to periodic behavior, respectively. Here, RC-para (no parameter) refers to the tipping point prediction directly through the RC extrapolation prediction, and the green dashed line illustrates the change in system parameters.}
	\label{F_KS}
	\vspace{-0.3cm}
\end{figure}

\subsubsection*{The Kuramoto-Sivashinsky System}
We next investigate the well-known Kuramoto-Sivashinsky (KS) system which is governed by a partial differential equation in \eqref{E_KS} and exhibits spatiotemporal chaos \cite{kuramoto1978diffusion,sivashinsky1980flame}. To validate the effectiveness of our method, we set $L = 2\pi$, $p(t) = 0.0056 \times \frac{t}{T\Delta t} + 0.076$, $T = 20,000$, $\Delta t = 0.02$, and $\omega = 10^{-4}$. We use a spatial discretization of $64$ points to generate the experimental time series data. Under these conditions, the system undergoes a transition from periodic dynamics to chaos \cite{akrivis2012computational}, with the bifurcation point occurring at approximately $p^\star \approx 0.08$. As shown in Fig.~\ref{fg:KSA}, our RCDyM method using the MFM effectively predicts this critical transition.  It is noted that the existing RC-para baseline methods \cite{kong2021machine,Patel2022UsingML,panahi2024adaptable} predict tipping points by feeding system parameters into the model and then forecasting future states.  However, in many practical situations, the parameter information is often unavailable. Thus, we use a modified baseline method, termed \textit{RC-para} ({\it no parameter}), which is trained exclusively on pre-tipping observational data without incorporating external parameters.   Consequently, the brown dashed lines in Fig.~\ref{fg:KSA} represent state predictions generated by this modified method.  Clearly, although the RC-para (no parameter) baseline effectively captures periodic patterns, it fails to predict tipping-point transitions when compared to our method under a fair and comparable configuration.

Furthermore, the spatial domain size $L$ is also a critical bifurcation parameter \cite{edson2019lyapunov}. In this case, the KS system exhibits highly complex chaotic behavior, necessitating an increased quantity of observational data to train the RC for learning the MLE of the original system. Here, we apply a stepwise decrease to $L$, as depicted by the green dashed line in the lower panel of Fig.~\ref{fg:KSB}, where each step contains $10,000$ data points. We then set $\Delta t = 0.25$, $p = 1$, and $\omega = 10^{-4}$ to generate the experimental time series data. Under these conditions, the system transits from chaos to periodic behavior, with the bifurcation occurring at approximately $L^\star \approx 33.7$. As shown in Fig.~\ref{fg:KSB}, the MLE measure reveals patterns, akin to those in Fig.~\ref{fg:bifurD}. Thus, by selecting an appropriate threshold $\epsilon$ (e.g., 0.06), we can effectively predict this critical transition. As illustrated by the brown dashed lines in Fig.~\ref{fg:KSB}, the RC-para (no parameter) baseline method again fails to anticipate the tipping point where the system reverts to periodic dynamics.

\subsection*{Robustness of the RCDyM Method}
\begin{figure*}[t]
	\centering
	\includegraphics[width=1.0\linewidth]{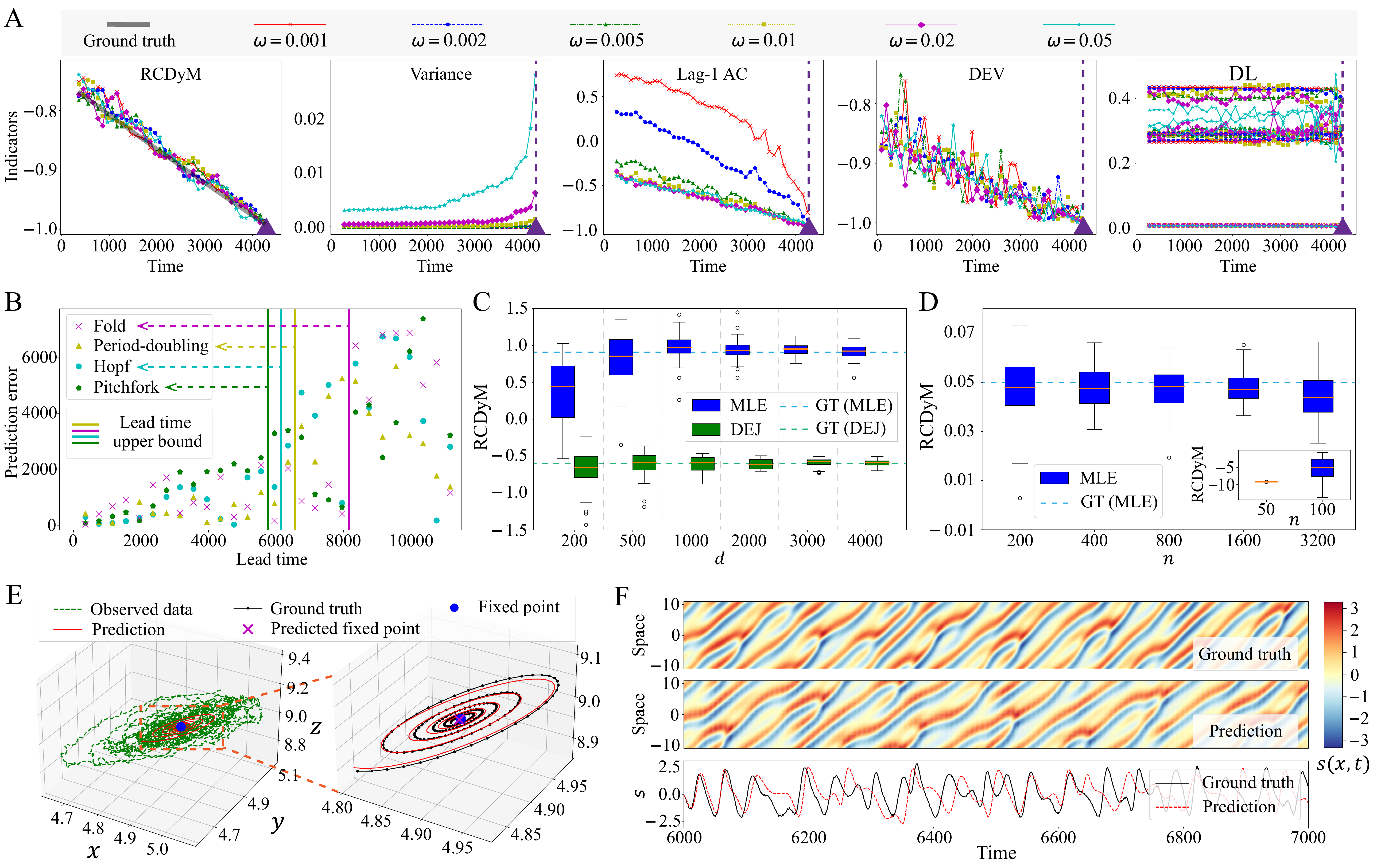}
	\subfigure{\label{fg:robustA}} \subfigure{\label{fg:robustB}}
	\subfigure{\label{fg:robustC}} \subfigure{\label{fg:robustD}}
	\subfigure{\label{fg:robustE}} \subfigure{\label{fg:robustF}}
	\vspace{-0.4cm}
	\caption{\textbf{Robustness demonstration of the RCDyM method.} ($A$) Predicting tipping points in the experiment of period-doubling bifurcation for different noise intensities. ($B$) Upper bound of the lead time for achieving ultra-early prediction of the tipping points for four bifurcation systems. ($C$) The estimations of the MLE and the DEJ in the Lorenz63 system for different window length $d$, compared to the ground-truth values of the DEJ at $-0.596$ ($p=10$) and the MLE at $0.91$ ($p=28$), respectively.  ($D$) The estimation of the MLE in the KS system for different $n$, with the ground-truth value of the MLE at 0.05 ($p=1$ and $L=22$). ($E$) Modeling dynamics near an equilibrium using the observational time series data from the Lorenz63 system. ($F$) Modeling the chaotic behavior of the KS system using the RC under an elevated noise level with $\omega=0.05$.}
	\label{F_robust}
	\vspace{-0.2cm}
\end{figure*}

\subsubsection*{Role of Noise}
The incorporation of the appropriate noise is crucial for accurately modeling local dynamics near the equilibrium. Without noise, the system remains perpetually in a state of equilibrium, rendering all tipping point prediction methods ineffective. Conversely, the excessive noise can obscure the intrinsic dynamics of the system, complicating the extraction of meaningful information from the observational time series data for the tipping point prediction.

To investigate the impact of noise on various methods, we first assess the robustness of discrete bifurcation systems via varying the noise strengths $\omega$. We employ the period-doubling bifurcation system as a test case to compare the performance of our RCDyM method against several baselines: Variance, lag-1 autocorrelation (Lag-1 AC) \cite{Scheffer2009EarlywarningSF}, dynamical eigenvalue (DEV) \cite{Grziwotz2023AnticipatingTO}, and deep learning (DL) \cite{Bury2021DeepLF}. As illustrated in Fig.~\ref{fg:robustA}, compared to Variance and DL indicators, our method exhibits the significant capability and the invariant scale property for accurately and robustly predicting critical transitions across various noise intensities. Additionally, compared to DEV and Lag-1 AC indicators, our method demonstrates the enhanced stability and effectively estimates the ground-truth DEJ of the original system. Notably, our method allows for a more precise estimation of the equilibrium by numerically solving the learned autonomous system \eqref{autoRC}, as opposed to simply selecting the last data point of a data window (which is a standard way used in the DEV method). Furthermore, regression analysis of the proposed RCDyM enables ultra-early prediction of tipping points, whereas conventional indicators such as variance and DL lack dynamical interpretability and well-defined thresholds, thus failing to provide precise prediction of when critical transitions occur. To further substantiate the noise robustness, we present experimental results for the other bifurcation systems in Supporting Information Section 6.A. 

\subsubsection*{Ultra-Early Prediction of Tipping Points}
By analyzing the trend of the dynamical measures, our method enables ultra-early prediction of tipping points. To further validate this perspective, we define the time interval between the ultra-early prediction time and the critical transition time as the lead time (see Fig.~\ref{fg:modelC}), and investigate the predictive accuracy of tipping points under varying lead times. As shown in Fig.~\ref{fg:robustB}, we present the relationship between lead time and prediction error of tipping points in four bifurcation systems, and provide the upper bound of lead time for achieving relatively accurate ultra-early prediction. From Fig.~\ref{fg:robustB}, it is evident that once the lead time falls below a certain threshold, the regression curve can be employed to accurately predict tipping points. This lower bound typically emerges at an early stage preceding the critical transition of the system, thereby validating the timeliness of our proposed method in predicting tipping points. In addition, see Supporting Information Section 5.C for more experimental analysis. 

\subsubsection*{Influence of Hyperparameters}
In this section, we examine the robustness and sensitivity of the proposed RCDyM method with respect to key hyperparameters, including the window length $d$ and those hyperparameters in RC. 

The selection of an appropriate window length $d$ is paramount for the accurate online prediction of tipping points. To assess the robustness of our method with respect to $d$, we estimate the DEJ and the MLE for the Lorenz63 system by changing the values of $d$. In these experiments, we assume that the system's parameters within a given window remain approximately constant. The results are displayed in Fig.~\ref{fg:robustC} as the box plots, where each estimate is derived from $50$ experimental trials. It is seen that the estimation performance of our method becomes increasingly stable as $d$ increases. Notably, even for small $d$, our method still accurately estimates the DEJ on average. We thus regard that this robustness arises from the method's ability to capture local dynamics that are critical to system stability.  Furthermore, since the MLE characterizes the global dynamics of the systems, larger $d$ values are required for the effective training of RC models. Our experiments show that the approach becomes reliable when $d$ exceeds $500$.

Then, we analyze the impact of RC hyperparameters on the performance of our method, focusing specifically on the effect of $n$, the reservoir network size (i.e., the number of nodes), on the MLE estimation for the KS system.   We perform $50$ trials for each value of $n$, with the results summarized in the box plots shown in Fig.~\ref{fg:robustD}. The experiments reveal that small $n$ values lead to the inaccurate MLE estimates, whereas increasing $n$ improves the accuracy up to a certain threshold. Beyond this point, further increases provide no substantial improvement and may even lead to performance degradation due to overfitting. This saturation point highlights that the reservoir size must be carefully chosen for the optimal performance. Furthermore, details on the selection strategies for hyperparameters, along with the hyperparameter configurations for all experiments, are provided in Supporting Information Section 4.

\subsubsection*{Modeling Capability in CDSs}
Another key advantage of our approach lies in its ability to leverage observational data for the accurate modeling of unknown CDSs, which is particularly important given the increasing prevalence of high-throughput data.  To illustrate this, we carry out the following experiments.

We first utilize the Lorenz63 system as a paradigm. To generate the time series data, as shown by the green dashed line in the left panel of Fig.~\ref{fg:robustE}, we set the system's parameters as: $\rho = 10,\, T=3000, \, \Delta t=0.005$, and $\omega = 0.001$. Despite the presence of the irregular motion near the equilibrium caused by the noise, our method effectively learns the local dynamics of the system. To validate this, we employ the trained RC model to perform multi-step extrapolation predictions, using the last observed data point as the initial value. The trajectory is then simulated using the system's dynamical equations with $\omega = 0$ as the GT. As demonstrated in the right panel of Fig.~\ref{fg:robustE}, our method shows a high degree of congruence with the ground-truth trajectory, accurately predicting both the position of the equilibrium and the dominant eigenvalue at that point. Furthermore, in practical applications, we assume that variations in environmental parameters within a given data window are typically minor, ensuring that the local dynamics remain largely unchanged. Our method remains capable of extracting dynamical measures under these constraints, as confirmed by the experiments presented in our work.

In addition, under increased noise levels, we explore the application of RC to model the KS system, estimating its MLE. The experimental time series data is generated following the configuration depicted in Fig.~\ref{fg:robustE}, with the first $6000$ data points used as the training set. Figure \ref{fg:robustF} illustrates the results of the extrapolation predictions over $1000$ time steps. Typically, successfully capturing the relatively long-term dynamics in chaotic systems is particularly challenging, especially in the presence of significant process noise, but our method is still capable of successfully capturing the relatively long-term dynamics of the system, producing a smooth, noise-free forecast while maintaining consistency with the MLE of the original system. 


\begin{figure*}[t]
	\centering
	\includegraphics[width=1.0 \textwidth]{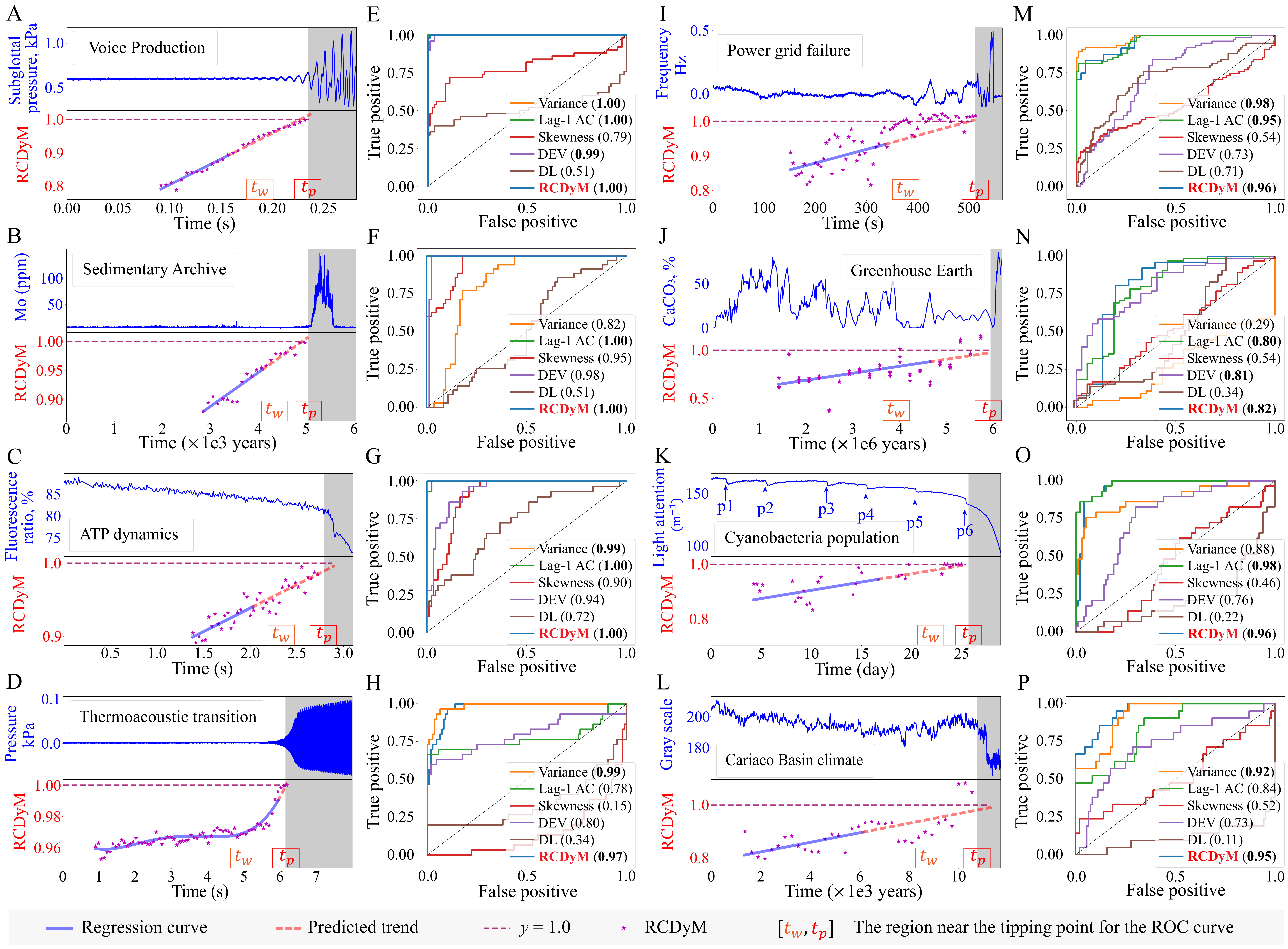}
	\subfigure{\label{fg:realA}}
	\subfigure{\label{fg:realB}}
	\subfigure{\label{fg:realC}}
	\subfigure{\label{fg:realD}}  
	\subfigure{\label{fg:realE}}
	\subfigure{\label{fg:realF}}
	\subfigure{\label{fg:realG}}
	\subfigure{\label{fg:realH}}
	\subfigure{\label{fg:realI}}
	\subfigure{\label{fg:realJ}}
	\subfigure{\label{fg:realK}}
	\subfigure{\label{fg:realL}}  
	\subfigure{\label{fg:realM}}
	\subfigure{\label{fg:realN}}
	\subfigure{\label{fg:realO}}
	\subfigure{\label{fg:realP}}
	\vspace{-0.5cm}
	\caption{\textbf{Performance of the RCDyM method in predicting tipping points across eight real-world datasets.} 
		($A$)-($H$) The experimental results for voice production, sedimentary archives, APT dynamics, and thermoacoustic transitions. ($I$)-($P$) The experimental results for power grid failure, greenhouse Earth conditions, cyanobacteria population dynamics, and the Cariaco Basin climate, respectively. Here, we denote the time interval corresponding to the top 30\% of indicators closest to the tipping point as $[t_w, t_p]$ and label this segment as 1, while all other regions are labeled 0. Thus, we obtain the ROC curve and AUC value, which are used to quantitatively evaluate the performance of our method in comparison to baseline approaches.}
	\label{F_real_data}
	\vspace{-0.2cm}
\end{figure*} 

\begin{figure*}[t]
	\centering
	\includegraphics[width=1.0 \textwidth]{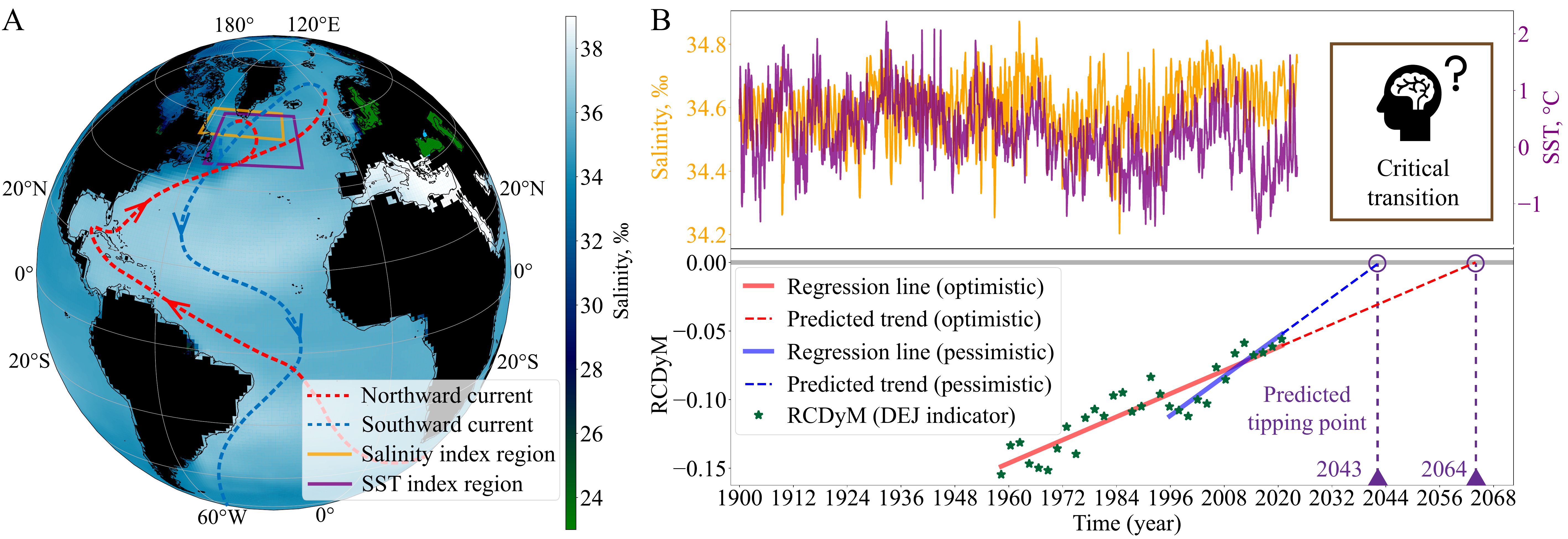}
	\subfigure{\label{fg:AMOCA}}
	\subfigure{\label{fg:AMOCB}}
	\vspace{-0.4cm}
	\caption{\textbf{Predicting the tipping point of the AMOC using the RCDyM method.} 
		($A$) The simplified schematic of the AMOC on an Earth map, with a color bar representing salinity levels in the ocean from $300$ meters depth to the surface, covering the first four months of 2024. 
		($B$) The upper panel displays the mean salinity and SST indices within the quadrilateral region highlighted in subfigure $A$, spanning the period from January 1900 to April 2024. The lower panel presents the EWS calculated using the continuous RCDyM method, where the red line represents the regression results using all the RCDyMs shown in the subfigure, while the blue line represents the regression results based on indicators from 1996 onward.}
	\label{F_real_AMOC}
	\vspace{-0.3cm}
\end{figure*} 

\subsection*{Applications to Real-World Datasets}
\subsubsection*{Critical Transitions in Eight Real-World Systems}
Here, we apply our framework of the RCDyM method to eight real-world datasets, each exhibiting critical transitions. These datasets include: a physical experiment on voice production during phonation onset \cite{Grziwotz2023AnticipatingTO}; sedimentary archives capturing periods of anoxia in the eastern Mediterranean region \cite{hennekam2020early}; cytosolic adenosine triphosphate (ATP) changes in live plants under progressively hypoxic conditions \cite{wagner2019multiparametric}; transitions towards thermoacoustic instability in a horizontal Rijke tube which is a prototypical thermoacoustic system \cite{pavithran2021effect}; the frequency of bus voltage prior to power grid collapse \cite{wscc1996western}; the abundance of calcium carbonate (CaCO$_3$) at the end of greenhouse Earth \cite{dakos2008slowing}; cyanobacteria microcosm experiment under light stress \cite{veraart2012recovery}; and Cariaco Basin 2000 Deglacial 14C and Grey Scale Data \cite{hughen2000cariaco}. Each of these datasets captures a critical transition triggered by environmental changes. Accurately predicting such transitions is notably challenging due to the nonlinearity and uncertainty inherent in real-world environments.

To facilitate data processing, we set the initial time of each experimental dataset to zero. We then apply our RCDyM method to predict tipping points within these systems, with the results illustrated in Fig.~\ref{F_real_data}. Both the discrete and continuous RCDyM methods provide significant early warnings prior to the occurrence of tipping points (see Supporting Information Section 5.F for continuous RCDyM results). In particular, Figures.~\ref{fg:realA} to \ref{fg:realD} demonstrate the robustness and the high performance of our method, likely due to the low levels of interference in the real-world data. However, in Figs.~\ref{fg:realI} to \ref{fg:realL}, the systems may be exposed to persistent, high-intensity disturbances, resulting in highly irregular observations. In such cases, existing methods often struggle to deliver stable predictions. Nonetheless, our approach outperforms SOTA baseline methods, with detailed comparisons provided in Supporting Information Section 6.B.

Moreover, for experiments depicted in Figs.~\ref{fg:realI} to \ref{fg:realL}, we improve the stability of our method by optimizing hyperparameters and excluding outlier data windows. For more unstable datasets, we recommend selecting a smaller hidden state dimension $n$ for the reservoir computing (RC) model and a larger regularization coefficient $\lambda$. Details on the selection strategies for hyperparameters, along with their configurations, can be found in Supporting Information Section 4. Additionally, outlier windows containing significant external disturbances can be omitted. For instance, in Fig.~\ref{fg:realK}, data windows involving manual interventions are excluded, which further enhances the stability of our method. The same procedures are applied to baseline methods to ensure fair comparisons.

In our previous discussions, the effectiveness of the proposed method was qualitatively demonstrated in aspects such as noise robustness, nonlinear processing capability, and ultra-early prediction. To complement this, we introduce a new quantitative metric to evaluate different approaches from the perspective of early-warning success rates for these eight real-world datasets. As shown in Fig.~\ref{F_real_data}, we select a time interval $[t_w, t_p]$ near the tipping point, which is defined as the ground truth of the warning region and labeled as $1$, while the remaining regions are labeled as $0$. Notably, a better prediction should exhibit values within $[t_w, t_p]$ that are closer to the critical threshold (e.g., RCDyM) or demonstrate larger magnitudes (e.g., variance) than those outside the warning region. This setup corresponds to a binary classification task for early-warning signals, allowing us to plot receiver operating characteristic (ROC) curves for different methods by varying the critical threshold (e.g., $\epsilon$ for RCDyM). As seen in Figs.~\ref{fg:realE}-\ref{fg:realH} and \ref{fg:realM}-\ref{fg:realP}, our method achieves the highest or near-highest area under the curve (AUC) across all eight experiments, demonstrating its superior and robust performance from the perspective of metric consistency in early-warning tasks.

Upon comparing AUC scores across different methods, we observe that the DL approach underperforms in out-of-distribution settings, indicating the need for further refinement. Similarly, the DEV method exhibits instability in more complex environments. In contrast, traditional statistical methods, such as variance and lag-1 AC, show relatively stable AUC performance. However, these methods occasionally fail, as seen with variance in the Greenhouse Earth experiment and lag-1 AC in the thermoacoustic transition experiment. And the skewness indicator is more commonly used in bistable or multistable systems and is widely observed in real-world scenarios; however, it still exhibits limitations in generality and robustness compared to the RCDyM approach. Additionally, our approach offers greater interpretability compared to statistical methods (e.g., variance and lag-1 AC) and DL methods, making it easier to select an appropriate threshold $\epsilon$ for ultra-early prediction of tipping points. In summary, we validate the effectiveness and robustness of our approach on complex real-world datasets through both qualitative and quantitative analyses.

\subsubsection*{Potential Collapse of the Atlantic Meridional Overturning Circulation}
The Atlantic Meridional Overturning Circulation (AMOC) is a crucial oceanic current system in the Atlantic Ocean, playing a significant role in regulating the global climate and supporting the marine ecosystems. As depicted by the dashed curve in Fig.~\ref{fg:AMOCA}, the AMOC transports warm, saline surface waters northward while moving colder, deeper waters southward. However, recent studies suggest that the AMOC is undergoing a critical transition from its current strong mode to a weaker mode \cite{boers2021observation,ben2023uncertainties}, which could have far-reaching consequences for the global climate and Earth system.

To evaluate the efficacy of our approach, we select two key observational variables related to the AMOC: the mean salinity in a specific region of the northern North Atlantic (54–62°N, 26–62°W) and the mean sea surface temperature (SST) of the subpolar gyre region (46–61°N, 20–55°W). As shown in the upper panel of Fig.~\ref{fg:AMOCB}, the salinity index reflects the mean salinity from $300$ meters depth to the ocean surface, while the SST variable is defined as the difference between the mean temperature within the SST index region and the global oceanic mean temperature.

We input SST and salinity as two-dimensional variables into the continuous-time RCDyM method to predict potential tipping points of the AMOC, with the corresponding results shown in the lower panel of Fig.~\ref{fg:AMOCB}. Here, we select appropriate RC hyperparameters to minimize the extrapolation prediction error $e_{\rm dyn}$ of Equation \eqref{E_edyn}, thereby improving the reliability of the tipping point predictions. The RCDyM shows a steady upward trend over time, approaching a critical threshold, which surprisingly aligns with the findings of previous studies \cite{boers2021observation,ditlevsen2023warning} (see Section 6.C of the Supporting Information for a further comparative analysis of RCDyM with these baseline methods). Notably, our method suggests that, if the current trend persists, a critical transition in the AMOC may occur around the year 2064, as indicated by the red line in the lower panel of Fig.~\ref{fg:AMOCB}. Furthermore, we observe an accelerated growth rate of the RCDyM after 1996, which can likely be attributed to the intensification of factors such as industrial activities, global warming, and ice sheet melting. Following this trend, the critical transition of the AMOC is expected to occur before the middle of this century, as indicated by the blue line in Fig.~\ref{fg:AMOCB}. It should be noted that the aforementioned conclusions were obtained under the condition that the dynamical index reached a threshold of zero. However, if the AMOC involves a noise-induced tipping process near the tipping point, the critical transition might occur earlier due to substantial perturbations. Therefore, it is crucial to continuously monitor the development of the AMOC and take effective measures to promptly reverse the trend toward its potential catastrophic collapse.

\section*{Concluding Remarks}
In this article, we introduced the RCDyM, a data-driven and machine-learning framework for predicting tipping points. The RCDyM framework employs a sliding window strategy to analyze local system dynamics from the observational time series data without any knowledge of the underlying CDSs. By examining the learned RC's dynamics, we derive several key dynamical measures, i.e., DEJ, MFM, and MFE, that serve as EWSs for critical transitions. The effectiveness and robustness of our method were validated through experiments on four bifurcation systems, three chaotic systems, and nine real-world datasets, demonstrating superior predictive performance and dynamical interpretability compared to SOTA baselines.   Notably, we provide a quantitative prediction and early warning analysis of a potential critical transition in an important climate system (AMOC), demonstrating that our framework enables ultra-early prediction of critical transitions through trend analysis of the RCDyMs.

Unlike conventional approaches, the RCDyM framework integrates RC modeling and dynamical systems theory, enabling more comprehensive extraction of three interpretable indicators from latent unknown CDSs directly from observed data. In addition to the commonly used DEJ measure, we extend our analysis by introducing two additional indicators: the MFM for periodic systems and the MLE for chaotic systems. Mathematically, these three measures are closely related. Specifically, DEJ and MFM can be regarded as degenerate forms of the MLE—at equilibrium points and limit cycles, respectively. Therefore, these three dynamical measures are fundamentally unified in their purpose: to quantify the local sensitivity and stability of dynamical systems in response to infinitesimal perturbations. These measures significantly enhance the versatility of our framework, allowing it to predict critical transitions in a wider variety of complex dynamics. In addition, in practical applications, the extrapolation predictive metrics (e.g., $e_{\rm dyn}$ in Eq.~\eqref{E_edyn}) can guide us in selecting appropriate RC hyperparameters for accurate prediction of tipping points, thus demonstrating the strong operability and practicality of our method.

Despite the promising results, there are three main limitations 
that warrant further investigation. 
First, in the real-world experiments, we primarily focus on using the DEJ measure for fixed-point systems, while the application of the MFM and the MLE in real-world contexts remains underexplored. Second, the RCDyM approach primarily focuses on bifurcation-induced tipping due to its stronger dynamical interpretability, while its efficacy for rate-induced tipping requires further validation and the development of a corresponding improved framework. Finally, this work primarily examines the RCDyM method in complex systems with relatively small numbers of variables. An important direction for future research is to extend our framework to networked dynamical systems by leveraging structural information, which will be a key focus of our ongoing work. We foresee broad applications of this framework for tipping point prediction across diverse real-world complex systems.

\section*{Materials and Methods}
In the first part of the Results section, we provided a comprehensive introduction to the RCDyM framework, and the main execution steps are shown in Algorithm \ref{alg:RCDI_framework}. In the following section, we provide additional implementation details of the RCDyM method and specify the dynamical models employed in the synthetic experiments of our work.
\begin{algorithm}[h]
	\caption{Main steps of the RCDyM framework}
	\label{alg:RCDI_framework}
	\begin{algorithmic}[1]
		\Require Observational time series data $\{s_1,s_2,...,s_T\}$.
		\State Set the values of $\epsilon$, $d$, $k$, and other RC hyperparameters; 
		\State Adjust the RC hyperparameters to minimize extrapolation prediction error $e_{dyn}$ in Eq.~\eqref{E_edyn}, then set $i=0$;
		\State Train RC using Eqs.~\eqref{contRC}-\eqref{Eloss} and data $\{s_{ik+1},\cdots,s_{ik+d}\}$ from $i$-th sliding window, and then derive the autonomous RC dynamics \eqref{autoRC};
		\State Calculate $i$-th dynamical indicator of RC Autonomous dynamics \eqref{autoRC}, including DEJ (Proposition \ref{P_DEJ}), MFM (Proposition \ref{P_MFM}), or MLE (Proposition \ref{P_MLE});
		\State If $(i+1)k+d \le T$, set $i = i+1$ and return to step 3; otherwise, proceed to the next step;
		\State Predict tipping points via measures form the RCDyM method. First, a critical transition is forecasted when the measures exceed the warning threshold. Then, the potential tipping point can be quantitatively predicted in advance by analyzing the trend of these measures;
		\Ensure Prediction of whether a tipping point is imminent.
	\end{algorithmic}
\end{algorithm}

\subsection*{Advantages in Critical Features}
By integrating the advanced machine learning technique with the dynamical theory, we introduce a novel and robust RCDyM method with the dynamical interpretability. Table~\ref{T_compare} presents a comprehensive comparison of key properties between the proposed method and the existing SOTA methods. The term ``Generality'' refers to the method's ability to predict any kind of CDSs solely using the time-series data, without additional information about the underlying systems such as time-varying parameters. ``Pre-tipping'' denotes that the method relies solely on data from the pre-tipping regime, eliminating the need for additional labeled data. ``Nonlinear'' indicates the model's capacity to handle potential nonlinear dynamics. ``Bifurcation types'' signifies the method’s ability to identify various prevalent bifurcation types. ``Periodic/Chaotic'' describes the capability to predict tipping points in systems exhibiting periodic or chaotic behavior. Finally, ``Ultra-early prediction'' refers to the capability of our RCDyM method to achieve quantitative prediction at a very early stage, well before the onset of critical transitions. And more detailed analyses about these features can be found in Supporting Information Section 2.E. The comparative analysis of these features reveals that the proposed method offers enhanced universality and a broader range of applications.

\begin{table*}[t]
	\centering
	\caption{\textbf{Comparison of the existing SOTA baselines and the proposed RCDyM method}.  Here, ``Generality'' refers to the applicability of the method to arbitrary systems without additional information about the time-varying parameters, ``Pre-tipping'' indicates whether the method requires data solely from the pre-tipping regime, ``Nonlinear'' indicates the model's capacity to handle potential nonlinear dynamics, ``Bifurcation types'' signifies the method’s ability to identify various prevalent bifurcation types, ``Periodic/Chaotic'' describes the capability to predict tipping points in systems exhibiting periodic or chaotic behavior, and ``Ultra-early prediction'' refers to the capability of our RCDyM method to achieve quantitative prediction at a very early stage, well before the onset of critical transitions.}
	\resizebox{1.0\linewidth}{!}{
		\begin{tabular}{cccccccc}
			\toprule
			Method & Reference & Generality  
			& Pre-tipping
			&Nonlinear
			&Bifurcation types &Periodic/Chaotic &Ultra-early prediction \\
			\midrule			
			Classical EWSs & Scheffer et al., 2009 \cite{Scheffer2009EarlywarningSF} &\CheckmarkBold 
			&\CheckmarkBold
			&N/A 
			&{\color{gray}\XSolidBrush}  &{\color{gray}\XSolidBrush} &{\color{gray}\XSolidBrush}\\
			DL & Bury et al., 2021 \cite{Bury2021DeepLF} 
			&\CheckmarkBold
			&{\color{gray}\XSolidBrush}
			&N/A 
			&\CheckmarkBold &{\color{gray}\XSolidBrush} &{\color{gray}\XSolidBrush} \\
			RC-para1 & Patel \& Ott, 2022 \cite{Patel2022UsingML} 
			&{\color{gray}\XSolidBrush}  
			&\CheckmarkBold
			&\CheckmarkBold 
			&{\color{gray}\XSolidBrush} &\CheckmarkBold &\CheckmarkBold\\
			DEV & Grziwotz et al., 2023 \cite{Grziwotz2023AnticipatingTO} &\CheckmarkBold 
			&\CheckmarkBold
			&{\color{gray}\XSolidBrush}   &\CheckmarkBold &{\color{gray}\XSolidBrush} &{\color{gray}\XSolidBrush} \\
			RC-para2 & Panahi et al., 2024 \cite{panahi2024adaptable} 
			&{\color{gray}\XSolidBrush} 
			&\CheckmarkBold
			&\CheckmarkBold   &{\color{gray}\XSolidBrush} &\CheckmarkBold &\CheckmarkBold\\ 
			RCDyM &This work &\CheckmarkBold 
			&\CheckmarkBold
			&\CheckmarkBold &\CheckmarkBold  &\CheckmarkBold &\CheckmarkBold \\
			\bottomrule
		\end{tabular}%
	}
	\label{T_compare}%
	\vspace{-0.4cm}
\end{table*}%

\subsection*{Multivariable Newton's method for finding equilibria} 
In computing the DEJ indicator using the RCDyM method (see Proposition \ref{P_DEJ}), we deviate from the baseline method (DEV), which involves randomly selecting a point for computation. Instead, we first solve for the equilibrium of the learned autonomous dynamics (see Eq.~\eqref{autoRC}) and then compute the DEJ index at this equilibrium. Specifically, we define the function $\bm{g}(r) = \gamma \{-\bm{r}(t) + \tanh[\tilde{\bm{A}}\bm{r}(t) + \tilde{\bm{b}}]\}$, and we use the final state of the controlled RC dynamics (see Eq.~\eqref{contRC}) as the initial guessed root $\bm{r}_0$ of $\bm{g}(r)$, serving as the starting point of the Newton iteration with the form: $\bm{r}_{i+1} = \bm{r}_{i} - \bm{J}_r^{-1} \bm{g}(\bm{r}_i)$. The iterations continue until the solution reaches a predefined precision threshold. Since our method models the local dynamics in the vicinity of equilibrium points, it typically exhibits superior convergence properties.

\begin{algorithm}[t]
	\caption{Automatic Detection of Periodicity Within Time Series Data}
	\label{alg:find_period}
	\begin{algorithmic}[1]
		\Require Data $\bm{s} = \{s_1,s_2,...,s_d\}$, initial value $t = T_{\text{min}}$, and threshold $\beta$.
		\State Let $\bm{s}_a = \{s_1,s_2,..., s_t\}$ and $\bm{s}_b = \{s_{t+1},s_{t+2},..., s_{2t}\}$. 
		\State We compute the correlation coefficient and error associated with $\bm{s}_a$ and $\bm{s}_b$, denoted as $c_t$ and $e_t$, respectively. 
		\State If $c_t > \beta$ and $e_t < 1 - \beta$, set the output period $T_\text{p}=t$; otherwise, increment $t$ by one and return to Step 1.
		\Ensure The period $T_\text{p}$ of the data $\bm{s}$.
	\end{algorithmic}
\end{algorithm}

\section*{Identifying the period of a periodic orbit} 
In computing the MFM indicator using the RCDyM method (see Proposition \ref{P_MFM}), we need to automatically recognize period. In practice, the presence of noise and varying sampling frequencies in observational time series data complicates the direct detection of periodic orbit periods through the simple measurement of the state difference between the start and end of the orbit. To facilitate automatic and robust detection of periodic orbit periods from the time series data, we employ autocorrelation coefficients combined with error analysis. The detailed detection procedure is outlined in Algorithm \ref{alg:find_period}.

\section*{Calculate MLE using QR decomposition} 
In computing the MLE indicator using the RCDyM method (see Proposition \ref{P_MLE}), we employ the QR decomposition method to calculate the MLE. We randomly generate a non-singular matrix $\bm{Q}_\text{ini}$ with dimension $n\times K$, and then orthonormalize it as $\bm{Q}_0$ by performing a QR decomposition $\bm{Q}_\text{ini} = \bm{Q}_0\bm{R}_0$. To ensure algorithm stability and efficiency, we choose an increasing sequence that tends to infinity, denoted as $0 = t_0 < t_1 < t_2 < \cdots$. For any $j \in \{0,1,2,...\}$, we have 
\begin{equation}
	\dot{\bm{Y}}_j (t) = \bm{J}_r[\bm{Y}_j(t)] \bm{Y}_j(t), \quad \bm{Y}_j(t_j) = \bm{Q}_j, \quad t_j \leq t \leq t_{j+1},
\end{equation}
and take the QR factorization $\bm{Y}_j(t_{j+1}) = \bm{Q}_{j+1} \bm{R}_{j+1}$.
Based on this, we estimate the $i$th Lyapunov exponent as follows:
$$
\lambda_i = \lim_{M\to \infty} \frac{1}{t_M}\log \left(\prod_{j=1}^M (\bm{R}_j)_{ii} \right), \quad 1\leq i\leq K.
$$
Therefore, we obtain the maximum Lyapunov exponent as $|\text{MLE}| = \lambda_1$. In practical applications, $M$ is typically chosen as a large finite value for approximate computation. Furthermore, when solely calculating the MLE, i.e., when $K$ equals 1, the aforementioned QR decomposition simplifies to the operation of vector norm calculation, thereby enhancing computational efficiency. 

\subsection*{Reservoir Computing Hyperparameter Selection Strategies}
To enhance the application of the proposed RCDyM method, beyond the adjustment of hyperparameters based on historical experience, we can further refine the selection of appropriate RC hyperparameters guided by the predictive performance of reservoir computing. This strategy leverages a fundamental principle: RC with superior dynamical prediction capabilities are expected to exhibit heightened tipping point forecasting abilities. 

Here, we employ the mean squared error (MSE) of one-step predictions as the metric to evaluate the predictive performance of RC. Specifically, when calculating the $i$-th RCDyM, we utilize the sliding window within the time interval $[t_i, t_{i+d}]$ as the training set, while use the data within the time interval $[t_{i+d}, t_{i+d+k}]$ as the extrapolated testing set. Subsequently, we calculate the mean of extrapolative predictions across all testing set to serve as the metric for assessing the dynamical forecasting performance of the RC under specific hyperparameters, denoted by 
\begin{equation}\label{E_edyn}
	e_{\rm dyn} = \frac{1}{n_d k} \sum_{i=0}^{n_d} \sum_{j=i+d}^{i+d+k} [\hat{s}(t_j)-s(t_j)]^2,
\end{equation}
where \(n_d\) represents the number of sliding windows, \(\hat{s}\) denotes the dynamical predictions of the RC, \(d\) signifies the window length, and \(k\) indicates the sliding step size. Finally, we prefer to select the RC hyperparameter that minimize $e_{\rm dyn}$, and the effectiveness of this strategy is validated in Supporting Information Section 4.

\subsection*{Definition of ``Ultra-Early Prediction''}
In our work, the dynamical measures generated by the proposed RCDyM method can not only serve as early-warning signals for critical transitions, but also enable ultra-early prediction of tipping points when trend-like patterns are present, as detailed in Definition \ref{ultra}.
	
\begin{definition} {\bf Ultra-Early Prediction of Tipping Points.}  
	Consider a dynamical system with a tipping point occurring at time \(t_p\). Let \(M(t)\) be a dynamical measure (e.g., DEJ) estimated via the RCDyM method using the sliding windows.
	
	1.  Early Trend Fitting: At a time \(t_l < t_p\), an approximating function \(P(t)\) is employed to fit the historical sequence of the measure \(\{M(t) : t < t_l\}\).
	
	2.  Extrapolation to Threshold: The predicted tipping time $\hat{t}_p$ is estimated by calculating the time at which $P(t)$ reaches the critical threshold $\tau_c$ (e.g., \(\tau_c = 0\) for the DEJ measure).
	
	This prediction is deemed ultra-early if: The measure \(M(t)\) exhibits a fittable trend pattern as the system approaches \(t_p\), which enables the simultaneous achievement of a significantly positive lead time (\(t_p - t_l \gg 0\)) and a sufficiently small prediction error (\(|t_p - \hat{t}_p| \to 0\)).
\end{definition} \label{ultra}

\subsection*{Dynamical Equations Underlying the Experimental Data} 
This section concisely outlines the dynamical equations underlying the simulation experiments presented in this work. We first consider four bifurcation systems: the corresponding dynamical equations, bifurcation types, and bifurcation points systematically summarized in Table \ref{T_bifur}. Next, we consider the discrete-time dynamical system, governed by the Logistic map, as:
\begin{equation}\label{logistic_map}
	s_{i+1}=ps_i(1-s_i) + \omega \zeta s_i,
\end{equation}
where $p$ represents the parameter related to the birth rate, and $\omega = 0.003$. Then we consider a more complex chaotic Lorenz63 system \cite{lorenz1963chaos}, governed by the following differential equations:
\begin{equation}\label{E_lorenz63}
	\begin{split}
		\dot{x} = 10 (y-x) + \omega \zeta x,\\
		\dot{y} = p x-y-xz + \omega \zeta y,\\
		\dot{z} = xy-8/3 z + \omega \zeta z,\\
	\end{split}
\end{equation}
where $p$ is a time-varying parameter and the state variables are $\bm{s} = \{x, y, z\}$. We finally consider the Kuramoto-Sivashinsky (KS) system, governed by the partial differential equation with periodic boundary conditions:
\begin{equation}\label{E_KS}
	\begin{aligned}
		\partial_t s(x,t) = &-s(x,t)\partial_x s(x,t) - \partial_{xx}s(x,t) \\
		& - p(t)\partial_{xxxx}s(x,t) + \omega \zeta s(x,t), \\
		s(x,t) =&\,\, s(x + L,t), \\
	\end{aligned}
\end{equation}
where $s(x,t)$ is a real-valued scalar function dependent on both space and time, with $x \in [-L/2, L/2]$ and $t \in [0, T\Delta t]$. 
Here, $L$ represents the spatial domain size, and $p(t)$ is a time-dependent parameter, functioning as the system's viscosity.  This KS system exhibits the spatiotemporal chaos, as reported in \cite{kuramoto1978diffusion,sivashinsky1980flame}.

\begin{table}[h]
	\centering
	\caption{\textbf{Dynamical equations and bifurcation values $p^\star$ of several bifurcation experiments.}}
	\resizebox{1.0\linewidth}{!}{
		\begin{tabular}{ccc}
			\toprule
			Bifurcation &$p^\star$ & Equation\\
			\midrule
			Fold &$1.82$ & $s_{i+1} = s_i \text{e}^{0.75-0.1s_i} - \frac{ps_i^2}{s_i^2+0.75^2} + \omega \zeta s_i$ \\
			\midrule 
			$\begin{aligned} &\text{\, Period-} \\ &\text{doubling} \end{aligned}$ &$0.37$ & $\begin{aligned}
				s_{1,i+1} &= 1 - ps_{1,i}^2 + s_{2,i} + \omega \zeta s_{1,i} \\
				&s_{2,i+1} = 0.3 s_{1,i} + \omega \zeta s_{2,i}
			\end{aligned}$ \\
			\midrule 
			Pitchfork &$1.19$ & $\dot{s}(t) = 0.5 + ps - s^3 + \omega \zeta s$ \\
			\midrule 
			Hopf &$0$ & $\begin{aligned}
				\dot{s}_1(t) = ps_1 - s_2 - s_1(s_1^2+s_2^2) + \omega \zeta s_1 \\
				\dot{s}_2(t) = s_1 + ps_2 - s_2(s_1^2+s_2^2) + \omega \zeta s_2
			\end{aligned}$ \\
			\bottomrule
		\end{tabular}
	}
	\label{T_bifur}
\end{table}

\section*{Data, Materials, and Software Availability} 
All simulated data used in this study were generated using programming simulations based on dynamical equations. The sea surface temperature (SST) data are publicly available at \url{https://www.metoffice.gov.uk/hadobs/hadisst}, and the ocean salinity data can be downloaded from \url{https://www.metoffice.gov.uk/hadobs/en4/}. Upon acceptance of the paper, all analysis code and datasets used in this study will be made publicly available on GitHub. 

\section*{ACKNOWLEDGMENTS}
X.L. is supported by the National Natural Science Foundation of China (No. 62506370), Innovation Research Foundation of (NUDT) and Hunan Provincial Natural Science Foundation of China (No. 2026JJ60049). Q.Z. is supported by the National Natural Science Foundation of China (Nos. 62406072 and 12171350) and by the STCSM (Nos. 23YF1402500, 21511100200, 22ZR1407300, and 22dz1200502). W.L. is supported by the National Natural Science Foundation of China (No. 11925103) and by the STCSM (Nos. 22JC1402500, 22JC1401402, and 2021SHZDZX0103).

\nocite{*}

\bibliography{apssamp}

@article{pathak2018model,
  title={Model-free prediction of large spatiotemporally chaotic systems from data: A reservoir computing approach},
  author={Pathak, Jaideep and Hunt, Brian and Girvan, Michelle and Lu, Zhixin and Ott, Edward},
  journal={Physical review letters},
  volume={120},
  number={2},
  pages={024102},
  year={2018},
  publisher={APS}
}

@article{zhu2019detecting,
  title={Detecting unstable periodic orbits based only on time series: When adaptive delayed feedback control meets reservoir computing},
  author={Zhu, Qunxi and Ma, Huanfei and Lin, Wei},
  journal={Chaos: An Interdisciplinary Journal of Nonlinear Science},
  volume={29},
  number={9},
  year={2019},
  publisher={AIP Publishing}
}

@article{lenton2019climate,
  title={Climate tipping points—too risky to bet against},
  author={Lenton, Timothy M and Rockstr{\"o}m, Johan and Gaffney, Owen and Rahmstorf, Stefan and Richardson, Katherine and Steffen, Will and Schellnhuber, Hans Joachim},
  journal={Nature},
  volume={575},
  number={7784},
  pages={592--595},
  year={2019},
  publisher={Nature Publishing Group UK London}
}

@article{clark2002role,
  title={The role of the thermohaline circulation in abrupt climate change},
  author={Clark, Peter U and Pisias, Nicklas G and Stocker, Thomas F and Weaver, Andrew J},
  journal={Nature},
  volume={415},
  number={6874},
  pages={863--869},
  year={2002},
  publisher={Nature Publishing Group UK London}
}

@article{van2014critical,
  title={Critical slowing down as early warning for the onset and termination of depression},
  author={van de Leemput, Ingrid A and Wichers, Marieke and Cramer, Ang{\'e}lique OJ and Borsboom, Denny and Tuerlinckx, Francis and Kuppens, Peter and Van Nes, Egbert H and Viechtbauer, Wolfgang and Giltay, Erik J and Aggen, Steven H and others},
  journal={Proceedings of the National Academy of Sciences},
  volume={111},
  number={1},
  pages={87--92},
  year={2014},
  publisher={National Acad Sciences}
}

@article{Grziwotz2023AnticipatingTO,
  title={Anticipating the occurrence and type of critical transitions},
  author={Florian Grziwotz and Chun‐Wei Chang and Vasilis Dakos and Egbert H. van Nes and Markus Schwarzl{\"a}nder and Oliver Kamps and Martin Hessler and Isao T. Tokuda and Arndt Telschow and Chih‐hao Hsieh},
  journal={Science Advances},
  year={2023},
  volume={9},
  url={https://api.semanticscholar.org/CorpusID:255501650}
}

@article{Scheffer2009EarlywarningSF,
  title={Early-warning signals for critical transitions},
  author={Marten Scheffer and Jordi Bascompte and William A. Buz Brock and Victor A. Brovkin and Stephen R. Carpenter and Vasilis Dakos and Hermann Held and Egbert H. van Nes and Max Rietkerk and George Sugihara},
  journal={Nature},
  year={2009},
  volume={461},
  pages={53-59},
  url={https://api.semanticscholar.org/CorpusID:4001553}
}

@article{Bury2021DeepLF,
  title={Deep learning for early warning signals of tipping points},
  author={Thomas M. Bury and R. I. Sujith and Induja Pavithran and Marten Scheffer and Timothy M. Lenton and Madhur Anand and Chris T. Bauch},
  journal={Proceedings of the National Academy of Sciences of the United States of America},
  year={2021},
  volume={118},
  url={https://api.semanticscholar.org/CorpusID:237583501}
}

@article{Patel2022UsingML,
  title={Using Machine Learning to Anticipate Tipping Points and Extrapolate to Post-Tipping Dynamics of Non-Stationary Dynamical Systems},
  author={Dhruvit Patel and Edward Ott},
  journal={Chaos},
  year={2022},
  volume={33 2},
  pages={
          023143
        },
  url={https://api.semanticscholar.org/CorpusID:250243662}
}

@article{kong2021machine,
  title={Machine learning prediction of critical transition and system collapse},
  author={Kong, Ling-Wei and Fan, Hua-Wei and Grebogi, Celso and Lai, Ying-Cheng},
  journal={Physical Review Research},
  volume={3},
  number={1},
  pages={013090},
  year={2021},
  publisher={APS}
}

@article{Dakos2008SlowingDA,
  title={Slowing down as an early warning signal for abrupt climate change},
  author={Vasilis Dakos and Marten Scheffer and Egbert H. van Nes and Victor A. Brovkin and Vladimir Petoukhov and Hermann Held},
  journal={Proceedings of the National Academy of Sciences},
  year={2008},
  volume={105},
  pages={14308 - 14312},
  url={https://api.semanticscholar.org/CorpusID:17303364}
}

@article{Carpenter2006RisingVA,
  title={Rising variance: a leading indicator of ecological transition.},
  author={Stephen R. Carpenter and William A. Buz Brock},
  journal={Ecology letters},
  year={2006},
  volume={9 3},
  pages={
          311-8
        },
  url={https://api.semanticscholar.org/CorpusID:40425736}
}

@article{Guttal2008ChangingSA,
  title={Changing skewness: an early warning signal of regime shifts in ecosystems.},
  author={Vishwesha Guttal and Ciriyam Jayaprakash},
  journal={Ecology letters},
  year={2008},
  volume={11 5},
  pages={
          450-60
        },
  url={https://api.semanticscholar.org/CorpusID:10038457}
}

@article{Hastings2018TransientPI,
  title={Transient phenomena in ecology},
  author={Alan Hastings and Karen C. Abbott and Kim Cuddington and Tessa B. Francis and Gabriel Gellner and Ying-Cheng Lai and Andrew Yu. Morozov and Sergei V. Petrovskii and Katherine Scranton and Mary Lou Zeeman},
  journal={Science},
  year={2018},
  volume={361},
  url={https://api.semanticscholar.org/CorpusID:52170587}
}

@article{Lesterhuis2017DynamicVS,
  title={Dynamic versus static biomarkers in cancer immune checkpoint blockade: unravelling complexity},
  author={W. Joost Lesterhuis and Anthony Bosco and Michael Millward and Michael Millward and Michael Small and Michael Small and Anna K. Nowak and Anna K. Nowak and Richard A. Lake},
  journal={Nature Reviews Drug Discovery},
  year={2017},
  volume={16},
  pages={264-272},
  url={https://api.semanticscholar.org/CorpusID:3099180}
}

@article{May2008ComplexSE,
  title={Complex systems: Ecology for bankers},
  author={Robert M. May and Simon A. Levin and George Sugihara},
  journal={Nature},
  year={2008},
  volume={451},
  pages={893-895},
  url={https://api.semanticscholar.org/CorpusID:205036062}
}

@article{centola2018experimental,
  title={Experimental evidence for tipping points in social convention},
  author={Centola, Damon and Becker, Joshua and Brackbill, Devon and Baronchelli, Andrea},
  journal={Science},
  volume={360},
  number={6393},
  pages={1116--1119},
  year={2018},
  publisher={American Association for the Advancement of Science}
}

@article{Dakos2012RobustnessOV,
  title={Robustness of variance and autocorrelation as indicators of critical slowing down.},
  author={Vasilis Dakos and Egbert H. van Nes and Paolo D’Odorico and Marten Scheffer},
  journal={Ecology},
  year={2012},
  volume={93 2},
  pages={
          264-71
        },
  url={https://api.semanticscholar.org/CorpusID:40497051}
}

@article{Wissel1984AUL,
  title={A universal law of the characteristic return time near thresholds},
  author={Christian Wissel},
  journal={Oecologia},
  year={1984},
  volume={65},
  pages={101-107},
  url={https://api.semanticscholar.org/CorpusID:5987443}
}

@article{vanNes2007SlowRF,
  title={Slow Recovery from Perturbations as a Generic Indicator of a Nearby Catastrophic Shift},
  author={Egbert H. van Nes and Marten Scheffer},
  journal={The American Naturalist},
  year={2007},
  volume={169},
  pages={738 - 747},
  url={https://api.semanticscholar.org/CorpusID:6916712}
}

@article{Tang2020IntroductionTF,
  title={Introduction to Focus Issue: When machine learning meets complex systems: Networks, chaos, and nonlinear dynamics.},
  author={Yang Tang and J{\"u}rgen Kurths and Wei Lin and Edward Ott and Ljupco Kocarev},
  journal={Chaos},
  year={2020},
  volume={30 6},
  pages={
          063151
        },
  url={https://api.semanticscholar.org/CorpusID:220307852}
}

@article{Lukoeviius2009ReservoirCA,
  title={Reservoir computing approaches to recurrent neural network training},
  author={Mantas Luko\v{s}evi\v{c}ius and Herbert Jaeger},
  journal={Comput. Sci. Rev.},
  year={2009},
  volume={3},
  pages={127-149},
  url={https://api.semanticscholar.org/CorpusID:554006}
}

@article{Jaeger2001TheechoST,
  title={The ``echo state'' approach to analysing and training recurrent neural networks-with an erratum note},
  author={Jaeger, Herbert},
  journal={Bonn, Germany: German National Research Center for Information Technology GMD Technical Report},
  volume={148},
  number={34},
  pages={13},
  year={2001},
  publisher={Bonn}
}

@article{Li2023TippingPD,
  title={Tipping Point Detection Using Reservoir Computing},
  author={X. Li and Qunxi Zhu and Chengli Zhao and Xuzhe Qian and Xue Zhang and Xiaojun Duan and Wei Lin},
  journal={Research},
  year={2023},
  volume={6},
  url={https://api.semanticscholar.org/CorpusID:258976377}
}

@article{Platt2022ASE,
  title={A Systematic Exploration of Reservoir Computing for Forecasting Complex Spatiotemporal Dynamics},
  author={Jason A. Platt and Stephen G. Penny and Timothy A. Smith and Tse-Chun Chen and Henry D. I. Abarbanel},
  journal={Neural networks : the official journal of the International Neural Network Society},
  year={2022},
  volume={153},
  pages={
          530-552
        },
  url={https://api.semanticscholar.org/CorpusID:246240543}
}

@article{Ascoli2022EdgeOC,
  title={Edge of Chaos Theory Resolves Smale Paradox},
  author={Alon Ascoli and Ahmet Samil Demirkol and Ronald Tetzlaff and Leon O. Chua},
  journal={IEEE Transactions on Circuits and Systems I: Regular Papers},
  year={2022},
  volume={69},
  pages={1252-1265},
  url={https://api.semanticscholar.org/CorpusID:246170967}
}

@article{Carroll2020DoRC,
  title={Do Reservoir Computers Work Best at the Edge of Chaos?},
  author={Thomas L. Carroll},
  journal={Chaos},
  year={2020},
  volume={30 12},
  pages={
          121109
        },
  url={https://api.semanticscholar.org/CorpusID:227247666}
}

@article{Gilpin2021ChaosAA,
  title={Chaos as an interpretable benchmark for forecasting and data-driven modelling},
  author={William Gilpin},
  journal={ArXiv},
  year={2021},
  volume={abs/2110.05266},
  url={https://api.semanticscholar.org/CorpusID:238583409}
}

@article{flynn2021multifunctionality,
  title={Multifunctionality in a reservoir computer},
  author={Flynn, Andrew and Tsachouridis, Vassilios A and Amann, Andreas},
  journal={Chaos: An Interdisciplinary Journal of Nonlinear Science},
  volume={31},
  number={1},
  year={2021},
  publisher={AIP Publishing}
}

@book{abarbanel2012analysis,
  title={Analysis of observed chaotic data},
  author={Abarbanel, Henry},
  year={2012},
  publisher={Springer Science \& Business Media}
}

@article{barone1977floquet,
  title={Floquet theory and applications},
  author={Barone, SR and Narcowich, MA and Narcowich, FJ},
  journal={Physical Review A},
  volume={15},
  number={3},
  pages={1109},
  year={1977},
  publisher={APS}
}

@article{lorenz1963chaos,
  title={Chaos in meteorological forecast},
  author={Lorenz, Edward},
  journal={Journal of the Atmospheric Sciences},
  volume={20},
  number={2},
  pages={130--141},
  year={1963}
}

@article{hennekam2020early,
  title={Early-warning signals for marine anoxic events},
  author={Hennekam, Rick and van der Bolt, Bregje and van Nes, Egbert H and de Lange, Gert J and Scheffer, Marten and Reichart, Gert-Jan},
  journal={Geophysical Research Letters},
  volume={47},
  number={20},
  pages={e2020GL089183},
  year={2020},
  publisher={Wiley Online Library}
}

@article{wagner2019multiparametric,
  title={Multiparametric real-time sensing of cytosolic physiology links hypoxia responses to mitochondrial electron transport},
  author={Wagner, Stephan and Steinbeck, Janina and Fuchs, Philippe and Lichtenauer, Sophie and Els{\"a}sser, Marlene and Schippers, Jos HM and Nietzel, Thomas and Ruberti, Cristina and Van Aken, Olivier and Meyer, Andreas J and others},
  journal={New Phytologist},
  volume={224},
  number={4},
  pages={1668--1684},
  year={2019},
  publisher={Wiley Online Library}
}

@article{pavithran2021effect,
  title={Effect of rate of change of parameter on early warning signals for critical transitions},
  author={Pavithran, Induja and Sujith, RI},
  journal={Chaos: An Interdisciplinary Journal of Nonlinear Science},
  volume={31},
  number={1},
  year={2021},
  publisher={AIP Publishing}
}

@article{veraart2012recovery,
  title={Recovery rates reflect distance to a tipping point in a living system},
  author={Veraart, Annelies J and Faassen, Elisabeth J and Dakos, Vasilis and Van Nes, Egbert H and L{\"u}rling, Miquel and Scheffer, Marten},
  journal={Nature},
  volume={481},
  number={7381},
  pages={357--359},
  year={2012},
  publisher={Nature Publishing Group UK London}
}

@article{wscc1996western,
  title={Western Systems Coordinating Council disturbance report For the power system outages that occurred on the western interconnection on July 2, 1996 and July 3, 1996},
  author={WSCC Operations Committee and others},
  journal={Western Systems Coordinating Council},
  year={1996}
}

@article{dakos2008slowing,
  title={Slowing down as an early warning signal for abrupt climate change},
  author={Dakos, Vasilis and Scheffer, Marten and Van Nes, Egbert H and Brovkin, Victor and Petoukhov, Vladimir and Held, Hermann},
  journal={Proceedings of the National Academy of Sciences},
  volume={105},
  number={38},
  pages={14308--14312},
  year={2008},
  publisher={National Acad Sciences}
}

@article{hughen2000cariaco,
  title={Cariaco Basin 2000 Deglacial 14C and Grey Scale Data},
  author={Hughen, KA and Southon, JR and Lehman, SJ and Overpeck, JT},
  journal={NOAA/NGDC Paleoclimatology Program, IGBP Pages/World Data Center A for Paleoclimatology, Data Contribution Series},
  volume={69},
  year={2000}
}

@article{liu2019detection,
  title={Detection for disease tipping points by landscape dynamic network biomarkers},
  author={Liu, Xiaoping and Chang, Xiao and Leng, Siyang and Tang, Hui and Aihara, Kazuyuki and Chen, Luonan},
  journal={National science review},
  volume={6},
  number={4},
  pages={775--785},
  year={2019},
  publisher={Oxford University Press}
}

@article{michel2022early,
  title={Early warning signal for a tipping point suggested by a millennial Atlantic Multidecadal Variability reconstruction},
  author={Michel, Simon LL and Swingedouw, Didier and Ortega, Pablo and Gastineau, Guillaume and Mignot, Juliette and McCarthy, Gerard and Khodri, Myriam},
  journal={Nature Communications},
  volume={13},
  number={1},
  pages={5176},
  year={2022},
  publisher={Nature Publishing Group UK London}
}

@article{qin2023tipping,
  title={Tipping point for pulsatile oscillations in dynamical networks},
  author={Qin, Bo-Wei and Lin, Wei and others},
  journal={Physical Review Research},
  volume={5},
  number={4},
  pages={043209},
  year={2023},
  publisher={APS}
}

@article{panahi2023rate,
  title={Rate-induced tipping in complex high-dimensional ecological networks},
  author={Panahi, Shirin and Do, Younghae and Hastings, Alan and Lai, Ying-Cheng},
  journal={Proceedings of the National Academy of Sciences},
  volume={120},
  number={51},
  pages={e2308820120},
  year={2023},
  publisher={National Acad Sciences}
}

@article{li2024higher,
  title={Higher-order Granger reservoir computing: simultaneously achieving scalable complex structures inference and accurate dynamics prediction},
  author={Li, Xin and Zhu, Qunxi and Zhao, Chengli and Duan, Xiaojun and Zhao, Bolin and Zhang, Xue and Ma, Huanfei and Sun, Jie and Lin, Wei},
  journal={Nature Communications},
  volume={15},
  number={1},
  pages={2506},
  year={2024},
  publisher={Nature Publishing Group UK London}
}

@article{zhu2023leveraging,
  title={Leveraging neural differential equations and adaptive delayed feedback to detect unstable periodic orbits based on irregularly sampled time series},
  author={Zhu, Qunxi and Li, Xin and Lin, Wei},
  journal={Chaos: An Interdisciplinary Journal of Nonlinear Science},
  volume={33},
  number={3},
  year={2023},
  publisher={AIP Publishing}
}

@article{lorenz1996predictability,
author = {Lorenz, Edward},
year = {1995},
month = {01},
pages = {},
title = {Predictability: A problem partly solved},
volume = {1},
isbn = {9780521848824},
journal = {ECMWF Seminar Proceedings I},
doi = {10.1017/CBO9780511617652.004}
}

@article{karimi2010extensive,
  title={Extensive chaos in the Lorenz-96 model},
  author={Karimi, Alireza and Paul, Mark R},
  journal={Chaos: An interdisciplinary journal of nonlinear science},
  volume={20},
  number={4},
  year={2010},
  publisher={AIP Publishing}
}

@article{kuramoto1978diffusion,
  title={Diffusion-induced chaos in reaction systems},
  author={Kuramoto, Yoshiki},
  journal={Progress of Theoretical Physics Supplement},
  volume={64},
  pages={346--367},
  year={1978},
  publisher={Oxford University Press}
}

@article{sivashinsky1980flame,
  title={On flame propagation under conditions of stoichiometry},
  author={Sivashinsky, Gregory I},
  journal={SIAM Journal on Applied Mathematics},
  volume={39},
  number={1},
  pages={67--82},
  year={1980},
  publisher={SIAM}
}

@article{pathak2017using,
  title={Using machine learning to replicate chaotic attractors and calculate Lyapunov exponents from data},
  author={Pathak, Jaideep and Lu, Zhixin and Hunt, Brian R and Girvan, Michelle and Ott, Edward},
  journal={Chaos: An Interdisciplinary Journal of Nonlinear Science},
  volume={27},
  number={12},
  year={2017},
  publisher={AIP Publishing}
}

@article{edson2019lyapunov,
  title={Lyapunov exponents of the Kuramoto--Sivashinsky PDE},
  author={Edson, Russell A and Bunder, Judith E and Mattner, Trent W and Roberts, Anthony J},
  journal={The ANZIAM Journal},
  volume={61},
  number={3},
  pages={270--285},
  year={2019},
  publisher={Cambridge University Press}
}

@article{akrivis2012computational,
  title={Computational study of the dispersively modified Kuramoto--Sivashinsky equation},
  author={Akrivis, Georgios and Papageorgiou, Demetrios T and Smyrlis, Y-S},
  journal={SIAM Journal on Scientific Computing},
  volume={34},
  number={2},
  pages={A792--A813},
  year={2012},
  publisher={SIAM}
}

@article{boers2021observation,
  title={Observation-based early-warning signals for a collapse of the Atlantic Meridional Overturning Circulation},
  author={Boers, Niklas},
  journal={Nature Climate Change},
  volume={11},
  number={8},
  pages={680--688},
  year={2021},
  publisher={Nature Publishing Group}
}

@article{ben2023uncertainties,
  title={Uncertainties in critical slowing down indicators of observation-based fingerprints of the Atlantic Overturning Circulation},
  author={Ben-Yami, Maya and Skiba, Vanessa and Bathiany, Sebastian and Boers, Niklas},
  journal={Nature Communications},
  volume={14},
  number={1},
  pages={8344},
  year={2023},
  publisher={Nature Publishing Group UK London}
}

@article{van2024physics,
  title={Physics-based early warning signal shows that AMOC is on tipping course},
  author={van Westen, Ren{\'e} M and Kliphuis, Michael and Dijkstra, Henk A},
  journal={Science advances},
  volume={10},
  number={6},
  pages={eadk1189},
  year={2024},
  publisher={American Association for the Advancement of Science}
}

@article{panahi2024adaptable,
  title={Adaptable reservoir computing: A paradigm for model-free data-driven prediction of critical transitions in nonlinear dynamical systems},
  author={Panahi, Shirin and Lai, Ying-Cheng},
  journal={Chaos: An Interdisciplinary Journal of Nonlinear Science},
  volume={34},
  number={5},
  year={2024},
  publisher={AIP Publishing}
}

@article{ditlevsen2023warning,
  title={Warning of a forthcoming collapse of the Atlantic meridional overturning circulation},
  author={Ditlevsen, Peter and Ditlevsen, Susanne},
  journal={Nature Communications},
  volume={14},
  number={1},
  pages={1--12},
  year={2023},
  publisher={Nature Publishing Group}
}

@CONTROL{REVTEX42Control}

@CONTROL{apsrev42Control,author="08",editor="1",pages="0",title="0",year="1"}

\end{document}